\documentclass{article}

\usepackage{microtype}
\usepackage{graphicx}
\usepackage{subcaption}
\usepackage{booktabs} 

\usepackage{hyperref}

\usepackage[accepted]{icml2026}

\usepackage{amsmath}
\usepackage{amssymb}
\usepackage{mathtools}
\usepackage{amsthm}

\usepackage[capitalize,noabbrev]{cleveref}

\theoremstyle{plain}

\theoremstyle{definition}

\theoremstyle{remark}

\DeclareMathOperator*{\argmin}{arg\,min}

\newcommand{\br}[1]{\left({#1}\right)}

\newcommand{\norm}[1]{\left\lVert#1\right\rVert}
\newcommand{\abs}[1]{\left\vert#1\right\vert}

\usepackage{multirow}
\usepackage[table]{xcolor}

\definecolor{biasgreen}{RGB}{0, 102, 0}
\definecolor{biasred}{RGB}{153, 0, 0}

\usepackage{tikz}
\usetikzlibrary{shapes.geometric, positioning, fit, calc}

\newcommand{\tikzsnowflake}[1][0.15]{%
  \tikz[scale=#1, line width=0.8pt, baseline=-0.5ex]{
    \foreach \angle in {0,60,...,300} {
      \draw[rotate=\angle] (0,0) -- (0,1)
        (0,0.5) -- (0.2,0.7)
        (0,0.5) -- (-0.2,0.7);
    }
  }%
}

\usepackage{bbm}

\usepackage[textsize=tiny]{todonotes}

\icmltitlerunning{Fairness-Aware Low-Rank Representation Fine-Tuning}

\begin{document}

\twocolumn[
  \icmltitle{Fairness-Aware Low-Rank Representation Fine-Tuning}



  \icmlsetsymbol{equal}{*}

  \begin{icmlauthorlist}
    \icmlauthor{Parameswaran Kamalaruban}{comp}
    \icmlauthor{Mark Anderson}{comp}
    \icmlauthor{Stuart Burrell}{comp}
    \icmlauthor{Maeve Madigan}{comp}
    \icmlauthor{Piotr Skalski}{comp}
    \icmlauthor{David Sutton}{comp}
  \end{icmlauthorlist}

  \icmlaffiliation{comp}{Risk and Security AI Lab, Visa Inc}

  \icmlcorrespondingauthor{Parameswaran Kamalaruban}{kaparame@visa.com}

  \icmlkeywords{Machine Learning, ICML}

  \vskip 0.3in
]



\printAffiliationsAndNotice{}  

\begin{abstract}
Pre-trained foundation models can be efficiently adapted for specific tasks using Low-Rank Adaptation (LoRA), but the fairness properties of these adapted classifiers remain underexplored. Existing fairness-aware fine-tuning methods assume that sensitive attribute labels are available alongside downstream task labels, which often fails in practice due to user consent limitations or privacy constraints. To address this gap, we investigate fairness-aware LoRA fine-tuning using separate datasets for downstream tasks and sensitive attributes. We introduce four fairness-aware LoRA strategies: sensitive unlearning, adversarial debiasing, orthogonality-based disentanglement, and entropy maximization. Through comprehensive experiments on standard algorithmic fairness datasets using an ImageNet pre-trained ViT-Base model, we evaluate these methods across multiple utility and fairness metrics. Our orthogonality-based disentanglement and entropy maximization approaches consistently outperform standard fine-tuning in both overall utility and fairness, while adversarial debiasing shows less consistent improvements and sensitive unlearning proves ineffective for classification tasks. However, fairness-aware methods underperform on certain metrics like subgroup-wise false-positive rate ratios, highlighting fundamental incompatibilities between fairness objectives. These findings demonstrate the potential of fairness-aware LoRA fine-tuning while revealing inherent challenges of simultaneously optimizing multiple fairness criteria in parameter-efficient adaptation.
\end{abstract}
\section{Introduction}
\label{sec:introduction}

Pre-trained foundation models have catalyzed remarkable advances across domains such as computer vision and natural language processing~\cite{awais2025foundation,zhao2023survey,zhou2024comprehensive}. Their ability to transfer learned representations to diverse downstream tasks has led to widespread adoption. However, these models also inherit biases present in their training data, potentially resulting in unfair or discriminatory downstream predictions~\cite{bommasani2021opportunities,ali2023evaluating}. As models grow larger and are fine-tuned on specialized datasets to achieve peak performance, addressing these inherent biases becomes critical, particularly for sensitive applications in law, healthcare, or finance. Recent developments in parameter-efficient fine-tuning (PEFT) techniques~\cite{han2024parameter}, such as Low-Rank Adaptation (LoRA)~\cite{hu2022lora}, enable efficient adaptation of massive pre-trained models by updating only a small subset of parameters. Despite its computational benefits and modular design, the fairness implications of LoRA and related PEFT methods remain insufficiently understood. 

In this work, we explore LoRA adapters for bias mitigation in pre-trained foundation models. We consider a practically relevant setting where we lack a joint dataset $\mathcal{D} = \{(x, y, g)\}$ containing input features $x$, target labels $y$, and sensitive attributes $g$. Instead, we assume access to two separate datasets: a downstream task dataset $\mathcal{D}^\mathrm{(task)} = \{(x, y)\}$ and a sensitive attribute dataset $\mathcal{D}^\mathrm{(sen)} = \{(x', g)\}$, where $x'$ is drawn from a distribution similar to $x$. This setting reflects practical constraints where sensitive attribute labels may be available only for a subset of data due to user consent limitations, or where privacy regulations prohibit maintaining joint datasets with both task and sensitive labels. Our objective is to fine-tune the frozen foundation model to maximize task performance while enforcing fairness by making predictions invariant to sensitive attributes.

Bias mitigation strategies are typically categorized as pre-processing~\cite{feldman2015certifying,calmon2017optimized,kamiran2012data}, in-processing~\cite{zhang2018mitigating,kamishima2012fairness,agarwal2018reductions,agarwal2019fair}, or post-processing~\cite{pleiss2017fairness,hardt2016equality,kamiran2012decision} approaches. We focus on in-processing methods that integrate fairness constraints directly into model fine-tuning. Traditional fairness regularization techniques~\cite{kamishima2012fairness,agarwal2019fair} typically require joint datasets $\mathcal{D} = \{(x, y, g)\}$ to impose fairness penalties and are thus inapplicable to our setting. Instead, we investigate fair representation learning approaches that decouple sensitive attribute influences from learned representations. While fairness regularization methods optimize specific group fairness metrics (e.g., equalized odds, demographic parity) that are often mutually incompatible~\cite{kim2020fact,berk2021fairness}, fair representation learning aims to remove sensitive information from representations themselves, naturally supporting our separate dataset scenario.

Despite growing interest in LoRA's fairness implications, most studies focus on fairness-unaware fine-tuning. Recent work~\cite{ding2024fairness} evaluates subgroup fairness properties of LoRA but limits evaluation to standard downstream fine-tuning. Similarly, approaches like FairLoRA~\cite{sukumaran2024fairlora} and FairTune~\cite{duttfairtune} integrate fairness objectives but require joint datasets $\mathcal{D} = \{(x, y, g)\}$. In contrast, we leverage separate datasets ${(x,y)}$ and ${(x',g)}$ to develop four fairness-aware LoRA strategies: (i) sensitive unlearning (\textsc{Unl}), (ii) adversarial debiasing via gradient reversal (\textsc{Adv}), (iii) orthogonality-based disentanglement that decorrelates task and sensitive subspaces (\textsc{Dis}), and (iv) entropy maximization that encourages maximal uncertainty in sensitive predictions (\textsc{Ent}).

\textbf{Contributions.} Our contributions are: 
\begin{enumerate} 
\item We introduce four fairness-aware LoRA fine-tuning methods applicable to our separate dataset setting. 
\item We comprehensively benchmark these methods against a fairness-unaware baseline (\textsc{Erm}) across multiple utility and fairness metrics. 
\item Through extensive experiments on standard algorithmic fairness datasets using an ImageNet pre-trained 86M ViT-Base model, we demonstrate that \textsc{Dis} and \textsc{Ent} consistently outperform \textsc{Erm} in both utility and fairness, while \textsc{Adv} shows improvements but less consistently. \textsc{Unl} proves ineffective for classification, often underperforming \textsc{Erm}. 
\item We empirically highlight incompatibilities between fairness metrics, observing that while fairness-aware methods generally improve performance, they underperform \textsc{Erm} specifically on subgroup-wise false-positive rate ratios, illustrating the challenges of optimizing multiple fairness objectives simultaneously. 
\end{enumerate}

\section{Related Work}
\label{sec:related-work}

\textbf{Algorithmic Fairness in Machine Learning.} Algorithmic fairness is a rapidly evolving field focused on ensuring equitable outcomes in AI decision-making systems. Various approaches have been proposed to measure and mitigate biases within algorithms~\cite{verma2018fairness,bellamy2019ai,weerts2023fairlearn,barocas2023fairness}. Bias mitigation strategies are generally categorized into pre-processing, in-processing, and post-processing methods. Pre-processing methods modify the training data to reduce bias before model training~\cite{feldman2015certifying,calmon2017optimized,kamiran2012data}, while post-processing techniques adjust model predictions to improve fairness after training~\cite{pleiss2017fairness,hardt2016equality,kamiran2012decision}. In contrast, in-processing approaches incorporate fairness constraints directly into the learning objective and have been extensively studied in classical machine learning models~\cite{zhang2018mitigating,kamishima2012fairness,agarwal2018reductions,agarwal2019fair}. For a more comprehensive review of fairness and bias mitigation strategies, we refer interested readers to recent surveys~\cite{mehrabi2021survey,caton2020fairness,hort2024bias,wan2023processing,ashurst2023fairness}.

\textbf{Fairness-Aware Fine-Tuning.} Recent studies have explored fairness-aware fine-tuning in deep learning. For instance, recent work~\cite{mao2023last} investigates last-layer fairness fine-tuning, where fairness constraints are introduced as regularization terms during the fine-tuning phase to mitigate bias in pre-trained deep neural networks. Methods such as FairLoRA~\cite{sukumaran2024fairlora} and FairTune~\cite{duttfairtune} have been proposed to integrate fairness considerations directly into the fine-tuning process. FairLoRA augments the downstream classification loss with a fairness regularization term based on specific fairness metrics, whereas FairTune adopts a bi-level optimization framework where an inner loop performs standard downstream fine-tuning with masked LoRA modules and an outer loop adjusts these masks to optimize a fairness objective on a validation set. However, these approaches typically require joint datasets $\mathcal{D} = \{(x, y, g)\}$.


\section{Problem Setup and Background}
\label{sec:problem-background}

\begin{figure*}[ht]
    \centering
    \begin{subfigure}[b]{0.22\linewidth}
        \centering
        \resizebox{0.9\linewidth}{!}{%
            \begin{tikzpicture}[
    node distance=6mm and 10mm,
    every node/.style={font=\sffamily},
    box/.style={
        rectangle, draw=blue!80!black, very thick,
        minimum width=2.2cm, minimum height=2.7cm, align=center
    },
    freeze/.style={
        box, fill=blue!10
    },
    trapezium down/.style={
        trapezium, trapezium stretches=true,
        trapezium left angle=120, trapezium right angle=120,
        draw=biasgreen!80!black, very thick, fill=biasgreen!20,
        minimum width=1.9cm, minimum height=0.62cm
    },
    trapezium up/.style={
        trapezium, trapezium stretches=true,
        trapezium left angle=60, trapezium right angle=60,
        draw=biasgreen!80!black, very thick, fill=biasgreen!20,
        minimum width=1.9cm, minimum height=0.62cm
    },
    labelnote/.style={inner sep=1pt, align=center},
    arrow/.style={-latex, very thick}
]

\def\udrop{4mm}

\node[trapezium down] (lora-top) {};
\node[trapezium up, below=2mm of lora-top] (lora-bottom) {};
\node[draw=none, inner sep=0pt, fit=(lora-top)(lora-bottom)] (lora-block) {};

\node[freeze, anchor=east] (back-bone) at ($ (lora-block.west) + (-1.2cm,0) $) {
    {\large \tikzsnowflake[0.30]} \\[0.4em]
    {\large $\theta^{(\mathrm{pre})}$}
}; 

\node[draw=none, inner sep=0pt, fit=(lora-block)(back-bone)] (full-rep) {};

\node[trapezium up, above=1.0cm of full-rep] (task-head) {};


\node[right=2mm of lora-top.north, above=3mm of lora-top.north, anchor=west] {\large $\theta^{(\mathrm{task})}$};
\node[left=2mm of task-head.west, anchor=east] {\large $w^{(\mathrm{task})}$};

\draw[very thick]
(back-bone.south) -- ++(0,-\udrop)               
-| ($ (lora-block.south) + (0,-\udrop) $)        
-- (lora-block.south);                           

\draw[very thick]
(back-bone.north) -- ++(0,\udrop)               
-| ($ (lora-block.north) + (0,\udrop) $)        
-- (lora-block.north);    

\draw[very thick] ($ (full-rep.north) + (0,\udrop) $) -- (task-head.south);

\draw[very thick] (task-head.north) -- ++(0,8mm) node[right=1mm, midway] {\large $\widehat{y}(x)$};

\draw[very thick] ($ (full-rep.south) + (0,-\udrop) $) -- ++(0,-8mm) node[right=1mm, midway] {\large $x \in \mathcal{D}^{(\mathrm{task})}$};

\end{tikzpicture}
        }
        \caption{\textsc{Erm} for downstream task}
        \label{fig:task-erm}
    \end{subfigure}
    \begin{subfigure}[b]{0.22\linewidth}
        \centering
        \resizebox{0.9\linewidth}{!}{%
            \begin{tikzpicture}[
    node distance=6mm and 10mm,
    every node/.style={font=\sffamily},
    box/.style={
        rectangle, draw=blue!80!black, very thick,
        minimum width=2.2cm, minimum height=2.7cm, align=center
    },
    freeze/.style={
        box, fill=blue!10
    },
    trapezium down/.style={
        trapezium, trapezium stretches=true,
        trapezium left angle=120, trapezium right angle=120,
        draw=biasred!80!black, very thick, fill=biasred!20,
        minimum width=1.9cm, minimum height=0.62cm
    },
    trapezium up/.style={
        trapezium, trapezium stretches=true,
        trapezium left angle=60, trapezium right angle=60,
        draw=biasred!80!black, very thick, fill=biasred!20,
        minimum width=1.9cm, minimum height=0.62cm
    },
    labelnote/.style={inner sep=1pt, align=center},
    arrow/.style={-latex, very thick}
]

\def\udrop{4mm}

\node[trapezium down] (lora-top) {};
\node[trapezium up, below=2mm of lora-top] (lora-bottom) {};
\node[draw=none, inner sep=0pt, fit=(lora-top)(lora-bottom)] (lora-block) {};

\node[freeze, anchor=east] (back-bone) at ($ (lora-block.west) + (-1.2cm,0) $) {
    {\large \tikzsnowflake[0.30]} \\[0.4em]
    {\large $\theta^{(\mathrm{pre})}$}
}; 

\node[draw=none, inner sep=0pt, fit=(lora-block)(back-bone)] (full-rep) {};

\node[trapezium up, above=1.0cm of full-rep] (task-head) {};


\node[right=2mm of lora-top.north, above=3mm of lora-top.north, anchor=west] {\large $\theta^{(\mathrm{sen})}$};
\node[left=2mm of task-head.west, anchor=east] {\large $w^{(\mathrm{sen})}$};

\draw[very thick]
(back-bone.south) -- ++(0,-\udrop)               
-| ($ (lora-block.south) + (0,-\udrop) $)        
-- (lora-block.south);                           

\draw[very thick]
(back-bone.north) -- ++(0,\udrop)               
-| ($ (lora-block.north) + (0,\udrop) $)        
-- (lora-block.north);    

\draw[very thick] ($ (full-rep.north) + (0,\udrop) $) -- (task-head.south);

\draw[very thick] (task-head.north) -- ++(0,8mm) node[right=1mm, midway] {\large $\widehat{g}(x')$};

\draw[very thick] ($ (full-rep.south) + (0,-\udrop) $) -- ++(0,-8mm) node[right=1mm, midway] {\large $x' \in \mathcal{D}^{(\mathrm{sen})}$};

\end{tikzpicture}
        }
        \caption{\textsc{Erm} for sensitive task}
        \label{fig:sen-erm}
    \end{subfigure}
    \begin{subfigure}[b]{0.36\linewidth}
        \centering
        \resizebox{0.9\linewidth}{!}{%
            \begin{tikzpicture}[
    node distance=6mm and 10mm,
    every node/.style={font=\sffamily},
    box/.style={
        rectangle, draw=blue!80!black, very thick,
        minimum width=2.2cm, minimum height=2.7cm, align=center
    },
    freeze/.style={
        box, fill=blue!10
    },
    trapezium down/.style={
        trapezium, trapezium stretches=true,
        trapezium left angle=120, trapezium right angle=120,
        draw=biasgreen!80!black, very thick, fill=biasgreen!20,
        minimum width=1.9cm, minimum height=0.62cm
    },
    trapezium up/.style={
        trapezium, trapezium stretches=true,
        trapezium left angle=60, trapezium right angle=60,
        draw=biasgreen!80!black, very thick, fill=biasgreen!20,
        minimum width=1.9cm, minimum height=0.62cm
    },
    trapezium down-freeze/.style={
        trapezium, trapezium stretches=true,
        trapezium left angle=120, trapezium right angle=120,
        draw=biasred!80!black, very thick, fill=biasred!10,
        minimum width=1.9cm, minimum height=0.62cm
    },
    trapezium up-freeze/.style={
        trapezium, trapezium stretches=true,
        trapezium left angle=60, trapezium right angle=60,
        draw=biasred!80!black, very thick, fill=biasred!10,
        minimum width=1.9cm, minimum height=0.62cm
    },
    labelnote/.style={inner sep=1pt, align=center},
    arrow/.style={-latex, very thick}
]

\def\udrop{4mm}

\node[trapezium down] (lora-top-target) {};
\node[trapezium up, below=2mm of lora-top-target] (lora-bottom-target) {};
\node[draw=none, inner sep=0pt, fit=(lora-top-target)(lora-bottom-target)] (lora-block-target) {};

\node[trapezium down-freeze, left=4.6cm of lora-top-target.west] (lora-top-sensitive) {\tikzsnowflake[0.20]};
\node[trapezium up-freeze, below=2mm of lora-top-sensitive] (lora-bottom-sensitive) {\tikzsnowflake[0.20]};
\node[draw=none, inner sep=0pt, fit=(lora-top-sensitive)(lora-bottom-sensitive)] (lora-block-sensitive) {};

\node[freeze, anchor=east] (back-bone) at ($ (lora-block-target.west) + (-1.2cm,0) $) {
    {\large \tikzsnowflake[0.30]} \\[0.4em]
    {\large $\theta^{(\mathrm{pre})}$}
}; 

\node[draw=none, inner sep=0pt, fit=(lora-block-target)(back-bone)(lora-block-sensitive)] (full-rep) {};

\node[trapezium up, above=1.0cm of full-rep] (task-head) {};

\draw[very thick]
(lora-block-sensitive.south) -- ++(0,-10mm)               
-| ($ (lora-block-target.south) + (0,-10mm) $)        
-- (lora-block-target.south);   

\draw[very thick]
(lora-block-sensitive.north) -- ++(0,10mm)               
-| ($ (lora-block-target.north) + (0,10mm) $)        
-- (lora-block-target.north);  

\draw[very thick] ($ (full-rep.south) + (0,-3.7mm) $) -- ++(0,3.7mm) {};

\draw[very thick] ($ (full-rep.south) + (0,-3.7mm) $) -- ++(0,-8mm) node[right=1mm, midway] {\large $x \in \mathcal{D}^{(\mathrm{task})}$};

\draw[very thick] (full-rep.north) -- (task-head.south);

\draw[very thick] (task-head.north) -- ++(0,8mm) node[right=1mm, midway] {\large $\widehat{y}(x)$};

\node[right=2mm of lora-top-target.north, above=3mm of lora-top-target.north, anchor=west] {\large $\theta^{(\mathrm{task})}$};
\node[left=2mm of lora-top-sensitive.north, above=3mm of lora-top-sensitive.north, anchor=east] {\large $- \lambda \cdot \theta^{(\mathrm{sen})}$};
\node[left=2mm of task-head.west, anchor=east] {\large $w^{(\mathrm{task})}$};

\end{tikzpicture}
        }
        \caption{Final stage of \textsc{Unl}}
        \label{fig:unl}
    \end{subfigure}
    \begin{subfigure}[b]{0.25\linewidth}
        \centering
        \resizebox{0.9\linewidth}{!}{%
            \begin{tikzpicture}[
    node distance=6mm and 10mm,
    every node/.style={font=\sffamily},
    box/.style={
        rectangle, draw=blue!80!black, very thick,
        minimum width=2.2cm, minimum height=2.7cm, align=center
    },
    freeze/.style={
        box, fill=blue!10
    },
    grl/.style={
        rectangle, draw=red!80!black, very thick, fill=red!20,
        minimum width=1.9cm, minimum height=0.25cm, align=center
    },
    trapezium down/.style={
        trapezium, trapezium stretches=true,
        trapezium left angle=120, trapezium right angle=120,
        draw=biasgreen!80!black, very thick, fill=biasgreen!20,
        minimum width=1.9cm, minimum height=0.62cm
    },
    trapezium up/.style={
        trapezium, trapezium stretches=true,
        trapezium left angle=60, trapezium right angle=60,
        draw=biasgreen!80!black, very thick, fill=biasgreen!20,
        minimum width=1.9cm, minimum height=0.62cm
    },
    trapezium down-freeze/.style={
        trapezium, trapezium stretches=true,
        trapezium left angle=120, trapezium right angle=120,
        draw=biasred!80!black, very thick, fill=biasred!10,
        minimum width=1.9cm, minimum height=0.62cm
    },
    trapezium up-freeze/.style={
        trapezium, trapezium stretches=true,
        trapezium left angle=60, trapezium right angle=60,
        draw=biasred!80!black, very thick, fill=biasred!10,
        minimum width=1.9cm, minimum height=0.62cm
    },
    labelnote/.style={inner sep=1pt, align=center},
    arrow/.style={-latex, very thick}
]

\def\udrop{4mm}

\node[trapezium down] (lora-top) {};
\node[trapezium up, below=2mm of lora-top] (lora-bottom) {};
\node[draw=none, inner sep=0pt, fit=(lora-top)(lora-bottom)] (lora-block) {};

\node[freeze, anchor=east] (back-bone) at ($ (lora-block.west) + (-1.2cm,0) $) {
    {\large \tikzsnowflake[0.30]} \\[0.4em]
    {\large $\theta^{(\mathrm{pre})}$}
}; 

\node[draw=none, inner sep=0pt, fit=(lora-block)(back-bone)] (full-rep) {};

\node[draw=none, inner sep=0pt, above=20mm of full-rep] (head-block) {};
\node[trapezium up-freeze, left=5mm of head-block] (sensitive-task-head) {\tikzsnowflake[0.20]};
\node[trapezium up, right=5mm of head-block] (target-task-head) {};

\draw[very thick]
(sensitive-task-head.south) -- ++(0,-8mm)               
-| ($ (target-task-head.south) + (0,-8mm) $)        
-- (target-task-head.south);  

\draw[very thick] (target-task-head.north) -- ++(0,8mm) node[right=1mm, midway] {\large $\widehat{y}(x)$};
\draw[very thick] (sensitive-task-head.north) -- ++(0,8mm) node[right=1mm, midway] {\large $\widehat{g}(x')$};

\node[grl, below=2mm of sensitive-task-head.south] (sensitive-grl) {};

\node[right=2mm of lora-top.north, above=3mm of lora-top.north,, anchor=west] {\large $\theta^{(\mathrm{task})}$};
\node[left=2mm of sensitive-task-head.west, anchor=east] {\large $w^{(\mathrm{sen})}$};
\node[right=2mm of target-task-head.east, anchor=west] {\large $w^{(\mathrm{task})}$};
\node[left=2mm of sensitive-grl.west, anchor=east] {\small $\mathrm{GRL}$};

\draw[very thick]
(back-bone.south) -- ++(0,-\udrop)               
-| ($ (lora-block.south) + (0,-\udrop) $)        
-- (lora-block.south);                           

\draw[very thick]
(back-bone.north) -- ++(0,\udrop)               
-| ($ (lora-block.north) + (0,\udrop) $)        
-- (lora-block.north);    

\draw[very thick] ($ (full-rep.north) + (0,\udrop) $) -- ($ (head-block.south) + (0, -11.1mm) $);

\draw[very thick] ($ (full-rep.south) + (0,-\udrop) $) -- ++(0,-8mm) node[left=1mm, midway] {\large $x' \in \mathcal{D}^{(\mathrm{sen})}$} node[right=1mm, midway] {\large $x \in \mathcal{D}^{(\mathrm{task})}$};

\end{tikzpicture}
        }
        \caption{Final stage of \textsc{Adv}}
        \label{fig:adv}
    \end{subfigure}
    \begin{subfigure}[b]{0.30\linewidth}
        \centering
        \resizebox{0.9\linewidth}{!}{%
            \begin{tikzpicture}[
    node distance=6mm and 10mm,
    box/.style={
        rectangle, draw=blue!80!black, very thick,
        minimum width=2.2cm, minimum height=2.7cm, align=center
    },
    freeze/.style={
        box, fill=blue!10
    },
    task-emb/.style={
        rectangle, draw=biasgreen!80!black, very thick, fill=biasgreen!20!blue!20,
        minimum width=1.9cm, minimum height=0.25cm, align=center
    },
    sen-emb/.style={
        rectangle, draw=biasred!80!black, very thick, fill=biasred!30!blue!15,
        minimum width=1.9cm, minimum height=0.25cm, align=center
    },
    trapezium down/.style={
        trapezium, trapezium stretches=true,
        trapezium left angle=120, trapezium right angle=120,
        draw=biasgreen!80!black, very thick, fill=biasgreen!20,
        minimum width=1.9cm, minimum height=0.62cm
    },
    trapezium up/.style={
        trapezium, trapezium stretches=true,
        trapezium left angle=60, trapezium right angle=60,
        draw=biasgreen!80!black, very thick, fill=biasgreen!20,
        minimum width=1.9cm, minimum height=0.62cm
    },
    trapezium down-freeze/.style={
        trapezium, trapezium stretches=true,
        trapezium left angle=120, trapezium right angle=120,
        draw=biasred!80!black, very thick, fill=biasred!10,
        minimum width=1.9cm, minimum height=0.62cm
    },
    trapezium up-freeze/.style={
        trapezium, trapezium stretches=true,
        trapezium left angle=60, trapezium right angle=60,
        draw=biasred!80!black, very thick, fill=biasred!10,
        minimum width=1.9cm, minimum height=0.62cm
    },
    labelnote/.style={inner sep=1pt, align=center},
    arrow/.style={-latex, very thick}
]

\def\udrop{4mm}

\node[trapezium down] (lora-top) {};
\node[trapezium up, below=2mm of lora-top] (lora-bottom) {};
\node[draw=none, inner sep=0pt, fit=(lora-top)(lora-bottom)] (lora-block) {};

\node[freeze, anchor=east] (back-bone) at ($ (lora-block.west) + (-1.2cm,0) $) {
    {\large \tikzsnowflake[0.30]} \\[0.4em]
    {\large $\theta^{(\mathrm{pre})}$}
}; 

\node[draw=none, inner sep=0pt, fit=(lora-block)(back-bone)] (full-rep) {};

\node[trapezium up, above=20mm of full-rep] (task-head) {};

\node[right=2mm of lora-top.north, above=3mm of lora-top.north, anchor=west] {\large $\theta^{(\mathrm{task})}$};
\node[left=2mm of task-head.west, anchor=east] {\large $w^{(\mathrm{task})}$};

\draw[very thick]
($ (back-bone.south) + (1mm, 0)$) -- ++(0,-\udrop)               
-| ($ (lora-block.south) + (0,-\udrop) $)        
-- (lora-block.south);                           

\draw[very thick]
($ (back-bone.north) + (1mm, 0)$) -- ++(0,\udrop)               
-| ($ (lora-block.north) + (0,\udrop) $)        
-- (lora-block.north);    

\draw[very thick] ($ (full-rep.north) + (0,\udrop) $) -- (task-head.south);

\draw[very thick] (task-head.north) -- ++(0,8mm) node[right=1mm, midway] {\large $\widehat{y}(x)$};

\draw[very thick] ($ (full-rep.south) + (0,-\udrop) $) -- ++(0,-8mm) node[right=1mm, midway] {\large $x \in \mathcal{D}^{(\mathrm{task})}$};

\node[trapezium down-freeze, left=4.6cm of lora-top-target.west] (lora-top-sensitive) {\tikzsnowflake[0.20]};
\node[trapezium up-freeze, below=2mm of lora-top-sensitive] (lora-bottom-sensitive) {\tikzsnowflake[0.20]};
\node[draw=none, inner sep=0pt, fit=(lora-top-sensitive)(lora-bottom-sensitive)] (lora-block-sensitive) {};
\node[draw=none, inner sep=0pt, fit=(lora-block-sensitive)(back-bone)] (full-rep-sensitive) {};
\draw[very thick]
($ (back-bone.south) + (-1mm, 0)$) -- ++(0,-\udrop)               
-| ($ (lora-block-sensitive.south) + (0,-\udrop) $)        
-- (lora-block-sensitive.south);
\draw[very thick]
($ (back-bone.north) + (-1mm, 0)$) -- ++(0,\udrop)               
-| ($ (lora-block-sensitive.north) + (0,\udrop) $)        
-- (lora-block-sensitive.north);
\draw[very thick] ($ (full-rep-sensitive.south) + (0,-\udrop) $) -- ++(0,-8mm) node[left=1mm, midway] {\large $x \in \mathcal{D}^{(\mathrm{task})}$};

\node[task-emb, above=8mm of full-rep.north] (task-emb-block) {};
\node[sen-emb, above=8mm of full-rep-sensitive.north] (sen-emb-block) {};

\draw[very thick] (sen-emb-block.south) -- ++(0,-4mm);

\draw[very thick]
(sen-emb-block.north) -- ++(0,2mm)               
-| ($ (task-emb-block.north) + (-2mm,2mm) $)        
-- ($ (task-emb-block.north) + (-2mm,0) $);

\node[labelnote, text width=2.4cm, above=5mm of sen-emb-block.north, anchor=west] {\large $R_\textsc{Dis}$};

\node[left=2mm of lora-top-sensitive.north, above=3mm of lora-top-sensitive.north, anchor=east] {\large $\theta^{(\mathrm{sen})}$};

\end{tikzpicture}
        }
        \caption{Final stage of \textsc{Dis}}
        \label{fig:dis}
    \end{subfigure}
    \begin{subfigure}[b]{0.25\linewidth}
        \centering
        \resizebox{0.9\linewidth}{!}{%
            \begin{tikzpicture}[
    node distance=6mm and 10mm,
    every node/.style={font=\sffamily},
    box/.style={
        rectangle, draw=blue!80!black, very thick,
        minimum width=2.2cm, minimum height=2.7cm, align=center
    },
    freeze/.style={
        box, fill=blue!10
    },
    grl/.style={
        rectangle, draw=red!80!black, very thick, fill=red!20,
        minimum width=1.9cm, minimum height=0.25cm, align=center
    },
    trapezium down/.style={
        trapezium, trapezium stretches=true,
        trapezium left angle=120, trapezium right angle=120,
        draw=biasgreen!80!black, very thick, fill=biasgreen!20,
        minimum width=1.9cm, minimum height=0.62cm
    },
    trapezium up/.style={
        trapezium, trapezium stretches=true,
        trapezium left angle=60, trapezium right angle=60,
        draw=biasgreen!80!black, very thick, fill=biasgreen!20,
        minimum width=1.9cm, minimum height=0.62cm
    },
    trapezium down-freeze/.style={
        trapezium, trapezium stretches=true,
        trapezium left angle=120, trapezium right angle=120,
        draw=biasred!80!black, very thick, fill=biasred!10,
        minimum width=1.9cm, minimum height=0.62cm
    },
    trapezium up-freeze/.style={
        trapezium, trapezium stretches=true,
        trapezium left angle=60, trapezium right angle=60,
        draw=biasred!80!black, very thick, fill=biasred!10,
        minimum width=1.9cm, minimum height=0.62cm
    },
    labelnote/.style={inner sep=1pt, align=center},
    arrow/.style={-latex, very thick}
]

\def\udrop{4mm}

\node[trapezium down] (lora-top) {};
\node[trapezium up, below=2mm of lora-top] (lora-bottom) {};
\node[draw=none, inner sep=0pt, fit=(lora-top)(lora-bottom)] (lora-block) {};

\node[freeze, anchor=east] (back-bone) at ($ (lora-block.west) + (-1.2cm,0) $) {
    {\large \tikzsnowflake[0.30]} \\[0.4em]
    {\large $\theta^{(\mathrm{pre})}$}
}; 

\node[draw=none, inner sep=0pt, fit=(lora-block)(back-bone)] (full-rep) {};

\node[draw=none, inner sep=0pt, above=20mm of full-rep] (head-block) {};
\node[trapezium up-freeze, left=5mm of head-block] (sensitive-task-head) {\tikzsnowflake[0.20]};
\node[trapezium up, right=5mm of head-block] (target-task-head) {};

\draw[very thick]
(sensitive-task-head.south) -- ++(0,-8mm)               
-| ($ (target-task-head.south) + (0,-8mm) $)        
-- (target-task-head.south);  

\draw[very thick] (target-task-head.north) -- ++(0,8mm) node[right=1mm, midway] {\large $\widehat{y}(x)$};
\draw[very thick] (sensitive-task-head.north) -- ++(0,8mm) node[right=1mm, midway] {\large $\widehat{g}(x)$};


\node[right=2mm of lora-top.north, above=3mm of lora-top.north, anchor=west] {\large $\theta^{(\mathrm{task})}$};
\node[left=2mm of sensitive-task-head.west, anchor=east] {\large $w^{(\mathrm{sen})}$};
\node[right=2mm of target-task-head.east, anchor=west] {\large $w^{(\mathrm{task})}$};

\draw[very thick]
(back-bone.south) -- ++(0,-\udrop)               
-| ($ (lora-block.south) + (0,-\udrop) $)        
-- (lora-block.south);                           

\draw[very thick]
(back-bone.north) -- ++(0,\udrop)               
-| ($ (lora-block.north) + (0,\udrop) $)        
-- (lora-block.north);    

\draw[very thick] ($ (full-rep.north) + (0,\udrop) $) -- ($ (head-block.south) + (0, -11.1mm) $);

\draw[very thick] ($ (full-rep.south) + (0,-\udrop) $) -- ++(0,-8mm) node[right=1mm, midway] {\large $x \in \mathcal{D}^{(\mathrm{task})}$};

\end{tikzpicture}
        }
        \caption{Final stage of \textsc{Ent}}
        \label{fig:ent}
    \end{subfigure}
    \caption{Empirical risk minimization baseline and fairness-aware fine-tuning methods.}
    \label{fig:fine-tune-all}
\end{figure*}

\textbf{Problem Setup.} We consider a multi-class classification problem with a sensitive attribute. The downstream task dataset is denoted as  $\mathcal{D}^{(\text{task})} = \{(x,y)\}$, where $x \in \mathcal{X}$ represents the input features and $y \in \mathcal{Y}$ is the target label. The sensitive task dataset is given by $\mathcal{D}^{(\text{sen})} = \{(x',g)\}$, with $g \in \mathcal{G}$ representing the sensitive attribute and the inputs drawn from a distribution similar to that of $\mathcal{D}^{(\text{task})}$. Given a pre-trained, frozen foundation model, our objective is to fine-tune it so that the resulting downstream predictor achieves high task performance while promoting fairness by making its predictions invariant to the sensitive attribute $g$.

\textbf{Utility and Fairness Metrics.} We assess model performance using a comprehensive set of utility and group fairness metrics~\cite{weerts2023fairlearn,hardt2016equality,chouldechova2017fair,dwork2012fairness}. Utility is measured through standard metrics such as accuracy (\texttt{ACC}), precision (\texttt{PPV}), recall (\texttt{TPR}), false positive rate (\texttt{FPR}), area under the receiver operating characteristic curve (\texttt{AUROC}), and area under the precision-recall curve (\texttt{AUPRC}). In addition to these, we evaluate group fairness with respect to the sensitive attribute $g$ (e.g., gender, race) by considering both difference and ratio metrics. For a utility metric $U$, let $U_g$ denote its value on subgroup $g$. The difference metric is $\Delta U = \max_{g, g' \in \mathcal{G}} \abs{U_g - U_{g'}}$ and the ratio metric is $U\text{-ratio} = \min_{g, g' \in \mathcal{G}} \frac{U_g}{U_{g'}}$. Smaller $\Delta U$ and larger $U\text{-ratio}$ indicate improved group fairness.

\textbf{Low-Rank Adaptation.} LoRA is a widely used PEFT method introduced by Hu et al.~\cite{hu2022lora} that adapts pre-trained models by learning low-dimensional updates to the weight matrices. In LoRA, given a pre-trained weight matrix $W_0 \in \mathbb{R}^{d \times k}$, the model update is parameterized as a low-rank decomposition: $W ~=~ W_0 + \Delta W ~=~ W_0 + B A^\top$, where $A \in \mathbb{R}^{d \times r}$ and $B \in \mathbb{R}^{r \times k}$, with rank $r \ll \min\br{d,k}$. The matrices are initialized such that $A$ is sampled from a Gaussian distribution, $A \sim \mathcal{N}(0, \sigma^2)$ for a small $\sigma$, and $B$ is set to the zero matrix, ensuring that the initial update $\Delta W$ is zero and the pre-trained model’s behavior is preserved. During training, $W_0$ remains frozen, and only $A$ and $B$ are updated. For transformer-based architectures, LoRA adapters are typically applied to the query and value matrices of the self-attention layers, while an additional task-specific head is attached to the last layer for supervised learning. This approach significantly reduces the number of trainable parameters while maintaining competitive performance on downstream tasks. Recently, \cite{zhang2025lori} observed that freezing the randomly initialized LoRA-A matrices during training reduces parameter interference when merging multiple task‑specific adapters; we adopt this design choice across all our fine-tuning methods.
\section{Fair Representation Fine-Tuning}
\label{sec:bias-mitigation}

In this section, we present a set of LoRA-based fine-tuning strategies for learning fair representations in classification tasks. Our goal is to adapt a pre-trained model such that the learned representations are \emph{predictive} of the target attribute while being \emph{invariant} to a given sensitive attribute (e.g., gender, race), thereby mitigating the influence of sensitive information on downstream predictions. 

We begin with a standard empirical risk minimization (\textsc{Erm}) baseline, which fine-tunes the model solely for target prediction without any fairness constraint. While \textsc{Erm} often yields strong task accuracy, it does not address representation bias and may inadvertently preserve or even amplify correlations between sensitive and target attributes.

To address this, we introduce four fairness-aware LoRA fine-tuning methods designed to reduce predictability of the sensitive attribute:
\begin{enumerate}
\item Unlearning-based debiasing (\textsc{Unl}), which adapts an unlearning approach from language model detoxification~\cite{zhang2023composing} to classification, explicitly removing sensitive attribute information via a learned sensitive adapter;
\item Adversarial debiasing (\textsc{Adv}), which builds on adversarial representation learning~\cite{ganin2016domain,adel2019one} to encourage target-relevant but sensitive-invariant features via a gradient reversal layer, with stabilization enhancements to mitigate typical adversarial training instabilities;
\item Orthogonality-based disentanglement (\textsc{Dis}), a novel regularization (inspired by disentangled representation learning~\cite{creager2019flexibly,locatello2019fairness}) that enforces orthogonality between target-task and sensitive-attribute subspaces in the representation, thereby decorrelating the two; and
\item Entropy maximization regularization (\textsc{Ent}), a stable alternative to adversarial objectives that directly maximizes the entropy of sensitive attribute predictions, pushing them toward maximal uncertainty~\cite{roy2019mitigating}.
\end{enumerate}

\textbf{Empirical risk minimization (\textsc{Erm}).} This baseline adapts a frozen pre-trained model $\theta^{(\text{pre})}$ to a downstream task by learning a task-specific LoRA adapter $\theta^{(\text{task})}_{\textsc{Erm}}$ and classification head $w^{(\text{task})}_{\textsc{Erm}}$. Training minimizes the cross-entropy loss on the downstream dataset $\mathcal{D}^{(\text{task})}$ with an additional norm regularization~\cite{zhang2023adalora} to stabilize LoRA parameters. The regularizer is $R_{\textsc{Norm}} (\theta) = \sum_i \lVert (A_i)^\top A_i - I \rVert_F^2 + \lVert (B_i)^\top B_i - I \rVert_F^2$, where $\theta = \{(A_i, B_i)\}$ are LoRA matrices. The training objective is:
\begin{align}
& (\theta^{(\text{task})}_{\textsc{Erm}}, w^{(\text{task})}_{\textsc{Erm}}) \gets \argmin_{\theta^{(\text{task})}, w^{(\text{task})}} \big[ \lambda_{\textsc{Norm}} \cdot R_{\textsc{Norm}} (\theta^{(\text{task})}) \nonumber \\
& \quad \quad \quad \quad + \ell_{\text{CE}} (\theta^{(\text{pre})} \oplus \theta^{(\text{task})}, w^{(\text{task})}; \mathcal{D}^{(\text{task})}) \big], \label{eq:erm-down-obj}
\end{align} 
where $\oplus$ denotes adding the LoRA adapter to the frozen backbone. The hyperparameter $\lambda_{\textsc{Norm}}$ controls the strength of the regularization.

\textbf{Unlearning-based debiasing (\textsc{Unl}).} This method removes sensitive attribute information from the frozen backbone before downstream training. We first learn a sensitive attribute adapter $\theta^{(\text{sen})}_{\textsc{Erm}}$ and classification head on the sensitive dataset $\mathcal{D}^{(\text{sen})}$ by minimizing the following objective:
\begin{align}
& (\theta^{(\text{sen})}_{\textsc{Erm}}, w^{(\text{sen})}_{\textsc{Erm}}) \gets \argmin_{\theta^{(\text{sen})}, w^{(\text{sen})}} \big[ \lambda_{\textsc{Norm}} \cdot R_{\textsc{Norm}} (\theta^{(\text{sen})}) \nonumber \\
& \quad \quad \quad \quad + \ell_{\text{CE}} (\theta^{(\text{pre})} \oplus \theta^{(\text{sen})}, w^{(\text{sen})}; \mathcal{D}^{(\text{sen})}) \big], \label{eq:erm-sen-obj}
\end{align}
with the pre-trained model $\theta^{(\text{pre})}$ remaining frozen. Once $\theta^{(\text{sen})}_{\textsc{Erm}}$ is trained, we \emph{unlearn} sensitive information by subtracting its scaled contribution from the frozen pre-trained weights: $\theta^{(\text{pre})} \ominus  \lambda_{\textsc{Unl}} \cdot \theta^{(\text{sen})}_{\textsc{Erm}}$, where $\ominus$ denotes negating either the LoRA-(A) or LoRA-(B) weights, and $\lambda_{\textsc{Unl}}$ controls the removal strength. Finally, we train a downstream task adapter $\theta^{(\text{task})}_{\textsc{Unl}}$ and its classification head $w^{(\text{task})}_{\textsc{Unl}}$ on $\mathcal{D}^{(\text{task})}$: 
\begin{align}
& (\theta^{(\text{task})}_{\textsc{Unl}}, w^{(\text{task})}_{\textsc{Unl}}) \gets \argmin_{\theta^{(\text{task})}, w^{(\text{task})}} \big[ \lambda_{\textsc{Norm}} \cdot R_{\textsc{Norm}} (\theta^{(\text{task})}) \nonumber \\
& + \ell_{\text{CE}} (\theta^{(\text{pre})} \ominus  \lambda_{\textsc{Unl}} \cdot \theta^{(\text{sen})}_{\textsc{Erm}} \oplus \theta^{(\text{task})}, w^{(\text{task})}; \mathcal{D}^{(\text{task})}) \big], \label{eq:unl-obj}
\end{align}
while keeping both $\theta^{(\text{pre})}$ and $\theta^{(\text{sen})}_{\textsc{Erm}}$ frozen.

\textbf{Adversarial debiasing (\textsc{Adv}).} This method learns task-relevant representations while suppressing sensitive attribute information via adversarial training. The process consists of three stages. We first pre-train a downstream task adapter $\theta^{(\text{task})}_{\textsc{Erm}}$ and its classification head $w^{(\text{task})}_{\textsc{Erm}}$ on $\mathcal{D}^{(\text{task})}$ via Eq.~\eqref{eq:erm-down-obj}. Next, using the frozen $\theta^{(\text{task})}_{\textsc{Erm}}$, we train a sensitive attribute classification head $w^{(\text{sen})}_{\textsc{Erm}}$ on $\mathcal{D}^{(\text{sen})}$ by minimizing $w^{(\text{sen})}_{\textsc{Erm}} \gets \argmin_{w^{(\text{sen})}} \ell_{\text{CE}} (\theta^{(\text{pre})} \oplus \theta^{(\text{task})}_{\textsc{Erm}}, w^{(\text{sen})}; \mathcal{D}^{(\text{sen})})$, while keeping both $\theta^{(\text{pre})}$ and $\theta^{(\text{task})}_{\textsc{Erm}}$ frozen. Finally, we adversarially fine-tune the downstream task adapter and the classification head---initialized from the first stage---by simultaneously minimizing the downstream task loss and maximizing the sensitive attribute prediction loss (via a gradient reversal layer~\cite{ganin2016domain} applied before $w^{(\text{sen})}_{\textsc{Erm}}$). The resulting objective is
\begin{align}
& (\theta^{(\text{task})}_\textsc{Adv}, w^{(\text{task})}_\textsc{Adv}) \gets \argmin_{\theta^{(\text{task})}, w^{(\text{task})}} \big[ \lambda_{\textsc{Norm}} \cdot R_{\textsc{Norm}} (\theta^{(\text{task})}) \nonumber \\
& \quad \quad - \lambda_{\textsc{Adv}} \cdot \ell_{\text{CE}} (\theta^{(\text{pre})} \oplus \theta^{(\text{task})}, w^{(\text{sen})}_{\textsc{Erm}}; \mathcal{D}^{(\text{sen})}) \nonumber \\
& \quad \quad + \ell_{\text{CE}} (\theta^{(\text{pre})} \oplus \theta^{(\text{task})}, w^{(\text{task})}; \mathcal{D}^{(\text{task})}) \big], \label{eq:adv-step-3}
\end{align}
where $\theta^{(\text{task})}$ and $w^{(\text{task})}$ are initialized from stage one, and $\theta^{(\text{pre})}$ and $w^{(\text{task})}_{\textsc{Erm}}$ remain frozen.

\begin{figure*}[ht]
    \centering
    \begin{subfigure}[b]{0.24\linewidth}
        \centering
        \includegraphics[width=0.9\linewidth]{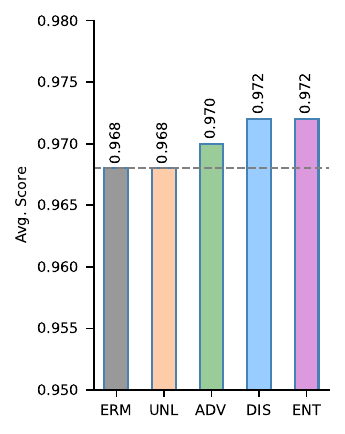}
        \caption{Overall AUROC Score ($\uparrow$)}
        \label{fig:avg-auroc}
    \end{subfigure}
    \begin{subfigure}[b]{0.24\linewidth}
        \centering
        \includegraphics[width=0.9\linewidth]{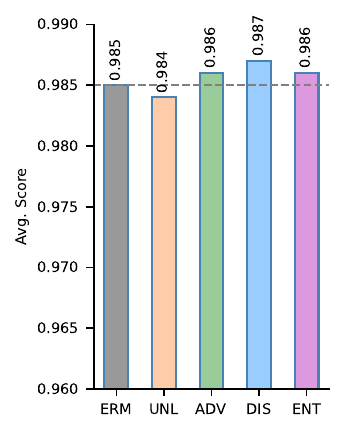}
        \caption{AUROC Ratio Score ($\uparrow$)}
        \label{fig:avg-auroc-ratio}
    \end{subfigure}
    \begin{subfigure}[b]{0.24\linewidth}
        \centering
        \includegraphics[width=0.9\linewidth]{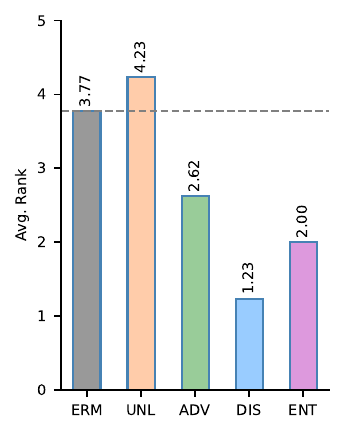}
        \caption{Overall AUROC Rank ($\downarrow$)}
        \label{fig:avg-auroc-rank}
    \end{subfigure}
    \begin{subfigure}[b]{0.24\linewidth}
        \centering
        \includegraphics[width=0.9\linewidth]{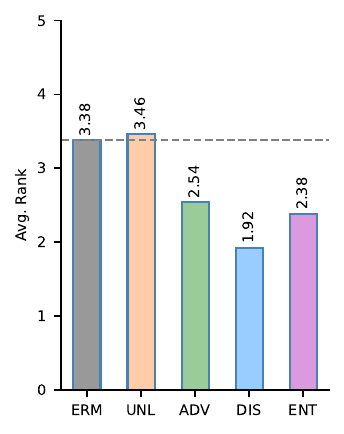}
        \caption{AUROC Ratio Rank ($\downarrow$)}
        \label{fig:avg-auroc-ratio-rank}
    \end{subfigure}
    \caption{Average overall AUROC and group-wise AUROC ratio (ratio of the best- to worst-performing subgroup) scores and ranks for each method, aggregated over all experiments.}
    \label{fig:auroc-average}
\end{figure*}

\begin{figure*}[ht]
    \centering
    \begin{subfigure}[b]{0.24\linewidth}
        \centering
        \includegraphics[width=0.9\linewidth]{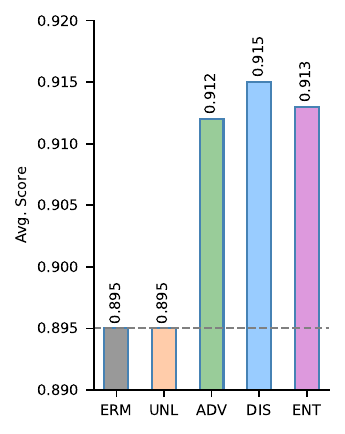}
        \caption{Overall AUPRC Score ($\uparrow$)}
        \label{fig:avg-auprc}
    \end{subfigure}
    \begin{subfigure}[b]{0.24\linewidth}
        \centering
        \includegraphics[width=0.9\linewidth]{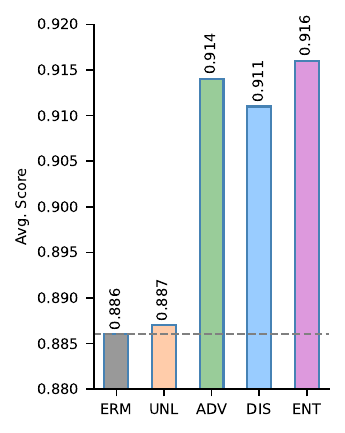}
        \caption{AUPRC Ratio Score ($\uparrow$)}
        \label{fig:avg-auprc-ratio}
    \end{subfigure}
    \begin{subfigure}[b]{0.24\linewidth}
        \centering
        \includegraphics[width=0.9\linewidth]{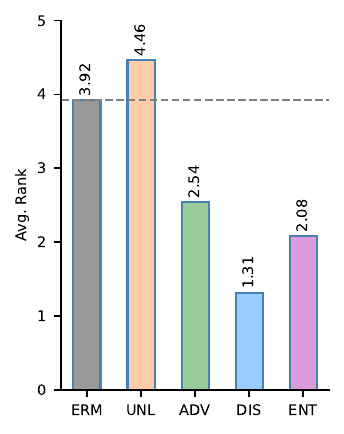}
        \caption{Overall AUPRC Rank ($\downarrow$)}
        \label{fig:avg-auprc-rank}
    \end{subfigure}
    \begin{subfigure}[b]{0.24\linewidth}
        \centering
        \includegraphics[width=0.9\linewidth]{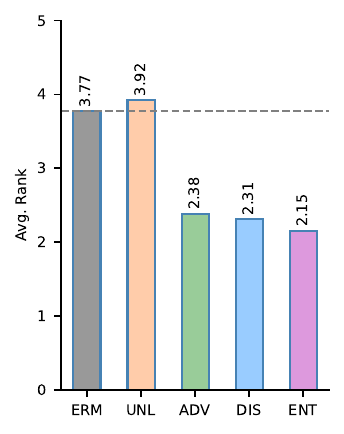}
        \caption{AUPRC Ratio Rank ($\downarrow$)}
        \label{fig:avg-auprc-ratio-rank}
    \end{subfigure}
    \caption{Average overall AUPRC and group-wise AUPRC ratio (ratio of the best- to worst-performing subgroup) scores and ranks for each method, aggregated over all experiments.}
    \label{fig:auprc-average}
\end{figure*}

\begin{table*}[ht]
\centering
\resizebox{0.7\linewidth}{!}{
\begin{tabular}{llccccc}
\toprule
\textbf{Metric} & \textbf{Method} & \multicolumn{2}{c}{\textbf{UTKFace}} & \multicolumn{2}{c}{\textbf{FairFace}} & \textbf{WaterBird} \\
\cmidrule(lr){3-4}
\cmidrule(lr){5-6}
\cmidrule(lr){7-7}
& & Age $\times$ Race & Gender $\times$ Race & Age $\times$ Race & Gender $\times$ Race & Type $\times$ Background \\
\midrule
\multirow{5}{*}{\texttt{ROC} ($\uparrow$)}
& \textsc{Erm} & \cellcolor{gray!20}0.942 ± 0.001 & \cellcolor{gray!20}0.985 ± 0.001 & \cellcolor{gray!20}0.915 ± 0.000 & \cellcolor{gray!20}0.984 ± 0.000 & \cellcolor{gray!20}0.981 ± 0.001 \\
& \textsc{Unl} & 0.945 ± 0.000 & 0.985 ± 0.001 & 0.913 ± 0.001 & 0.981 ± 0.000 & 0.982 ± 0.003 \\
& \textsc{Adv} & \textcolor{biasgreen}{0.950 ± 0.000} & 0.985 ± 0.001 & 0.916 ± 0.001 & 0.986 ± 0.000 & 0.977 ± 0.002 \\
& \textsc{Dis} & \textcolor{biasgreen}{\textbf{0.951 ± 0.001}} & 0.986 ± 0.001 & 0.919 ± 0.000 & \textbf{0.987 ± 0.000} & 0.986 ± 0.002 \\
& \textsc{Ent} & 0.947 ± 0.007 & \textbf{0.988 ± 0.001} & \textcolor{biasgreen}{\textbf{0.922 ± 0.001}} & \textbf{0.987 ± 0.000} & \textcolor{biasgreen}{\textbf{0.994 ± 0.001}} \\
\midrule
\multirow{5}{*}{$\Delta$\texttt{ROC} ($\downarrow$)}
& \textsc{Erm} & \cellcolor{gray!20}0.048 ± 0.006 & \cellcolor{gray!20}0.031 ± 0.002 & \cellcolor{gray!20}0.019 ± 0.001 & \cellcolor{gray!20}0.036 ± 0.004 & \cellcolor{gray!20}\textbf{0.004 ± 0.001} \\
& \textsc{Unl} & 0.049 ± 0.007 & \textbf{0.026 ± 0.002} & 0.023 ± 0.002 & 0.037 ± 0.003 & 0.007 ± 0.002 \\
& \textsc{Adv} & \textcolor{biasgreen}{\textbf{0.038 ± 0.002}} & \textbf{0.026 ± 0.002} & 0.020 ± 0.000 & 0.031 ± 0.003 & 0.007 ± 0.003 \\
& \textsc{Dis} & \textcolor{biasgreen}{0.042 ± 0.006} & 0.028 ± 0.001 & \textbf{0.018 ± 0.001} & \textcolor{biasgreen}{0.029 ± 0.003} & 0.005 ± 0.001 \\
& \textsc{Ent} & 0.046 ± 0.008 & 0.027 ± 0.001 & \textbf{0.018 ± 0.001} & \textcolor{biasgreen}{\textbf{0.028 ± 0.002}} & \textbf{0.004 ± 0.001} \\
\midrule
\multirow{5}{*}{\texttt{ROC} ratio ($\uparrow$)}
& \textsc{Erm} & \cellcolor{gray!20}0.950 ± 0.006 & \cellcolor{gray!20}0.968 ± 0.002 & \cellcolor{gray!20}0.979 ± 0.001 & \cellcolor{gray!20}0.964 ± 0.004 & \cellcolor{gray!20}\textbf{0.996 ± 0.001} \\
& \textsc{Unl} & 0.948 ± 0.007 & \textcolor{biasgreen}{\textbf{0.974 ± 0.002}} & 0.975 ± 0.002 & 0.962 ± 0.003 & 0.993 ± 0.002 \\
& \textsc{Adv} & \textcolor{biasgreen}{\textbf{0.960 ± 0.002}} & 0.973 ± 0.002 & 0.978 ± 0.001 & 0.969 ± 0.003 & 0.993 ± 0.003 \\
& \textsc{Dis} & \textcolor{biasgreen}{0.956 ± 0.006} & 0.972 ± 0.002 & \textbf{0.981 ± 0.001} & \textcolor{biasgreen}{0.971 ± 0.003} & 0.995 ± 0.001 \\
& \textsc{Ent} & 0.953 ± 0.008 & 0.973 ± 0.001 & \textbf{0.981 ± 0.001} & \textcolor{biasgreen}{\textbf{0.972 ± 0.002}} & \textbf{0.996 ± 0.001} \\
\midrule
\midrule
\multirow{5}{*}{\texttt{PR} ($\uparrow$)}
& \textsc{Erm} & \cellcolor{gray!20}0.676 ± 0.010 & \cellcolor{gray!20}0.978 ± 0.002 & \cellcolor{gray!20}0.558 ± 0.003 & \cellcolor{gray!20}0.982 ± 0.000 & \cellcolor{gray!20}0.951 ± 0.004 \\
& \textsc{Unl} & \textcolor{biasgreen}{0.707 ± 0.013} & 0.978 ± 0.003 & 0.555 ± 0.004 & 0.979 ± 0.000 & \textcolor{biasgreen}{0.961 ± 0.003} \\
& \textsc{Adv} & \textcolor{biasgreen}{0.814 ± 0.002} & 0.979 ± 0.001 & \textcolor{biasgreen}{0.575 ± 0.000} & 0.984 ± 0.000 & 0.947 ± 0.004 \\
& \textsc{Dis} & \textcolor{biasgreen}{0.815 ± 0.014} & 0.981 ± 0.002 & \textcolor{biasgreen}{0.583 ± 0.004} & \textbf{0.986 ± 0.000} & \textcolor{biasgreen}{0.962 ± 0.004} \\
& \textsc{Ent} & \textcolor{biasgreen}{\textbf{0.822 ± 0.006}} & \textbf{0.983 ± 0.002} & \textcolor{biasgreen}{\textbf{0.594 ± 0.002}} & 0.984 ± 0.000 & \textcolor{biasgreen}{\textbf{0.980 ± 0.002}} \\
\midrule
\multirow{5}{*}{$\Delta$\texttt{PR} ($\downarrow$)}
& \textsc{Erm} & \cellcolor{gray!20}0.237 ± 0.017 & \cellcolor{gray!20}0.039 ± 0.010 & \cellcolor{gray!20}0.099 ± 0.006 & \cellcolor{gray!20}0.031 ± 0.004 & \cellcolor{gray!20}0.052 ± 0.017 \\
& \textsc{Unl} & \textcolor{biasgreen}{0.190 ± 0.010} & 0.037 ± 0.011 & \textcolor{biasred}{0.107 ± 0.006} & 0.033 ± 0.002 & \textcolor{biasgreen}{0.035 ± 0.012} \\
& \textsc{Adv} & \textcolor{biasgreen}{\textbf{0.105 ± 0.009}} & \textcolor{biasgreen}{0.030 ± 0.004} & 0.104 ± 0.006 & 0.027 ± 0.004 & \textcolor{biasred}{0.067 ± 0.019} \\
& \textsc{Dis} & \textcolor{biasgreen}{0.164 ± 0.046} & \textcolor{biasgreen}{0.031 ± 0.008} & \textcolor{biasred}{0.107 ± 0.005} & \textcolor{biasgreen}{0.025 ± 0.003} & \textcolor{biasgreen}{0.031 ± 0.012} \\
& \textsc{Ent} & \textcolor{biasgreen}{0.144 ± 0.038} & \textcolor{biasgreen}{\textbf{0.028 ± 0.002}} & \textcolor{biasgreen}{\textbf{0.079 ± 0.019}} & \textcolor{biasgreen}{\textbf{0.025 ± 0.002}} & \textcolor{biasgreen}{\textbf{0.021 ± 0.014}} \\
\midrule
\multirow{5}{*}{\texttt{PR} ratio ($\uparrow$)}
& \textsc{Erm} & \cellcolor{gray!20}0.736 ± 0.015 & \cellcolor{gray!20}0.960 ± 0.010 & \cellcolor{gray!20}0.835 ± 0.009 & \cellcolor{gray!20}0.969 ± 0.004 & \cellcolor{gray!20}0.947 ± 0.018 \\
& \textsc{Unl} & \textcolor{biasgreen}{0.768 ± 0.007} & 0.962 ± 0.012 & \textcolor{biasred}{0.825 ± 0.007} & 0.967 ± 0.002 & \textcolor{biasgreen}{0.964 ± 0.012} \\
& \textsc{Adv} & \textcolor{biasgreen}{\textbf{0.881 ± 0.008}} & \textcolor{biasgreen}{0.970 ± 0.004} & 0.834 ± 0.009 & 0.973 ± 0.004 & \textcolor{biasred}{0.931 ± 0.019} \\
& \textsc{Dis} & \textcolor{biasgreen}{0.817 ± 0.050} & \textcolor{biasgreen}{0.969 ± 0.008} & 0.831 ± 0.007 & \textcolor{biasgreen}{0.975 ± 0.003} & \textcolor{biasgreen}{0.968 ± 0.012} \\
& \textsc{Ent} & \textcolor{biasgreen}{0.840 ± 0.040} & \textcolor{biasgreen}{\textbf{0.972 ± 0.002}} & \textcolor{biasgreen}{\textbf{0.875 ± 0.029}} & \textcolor{biasgreen}{\textbf{0.975 ± 0.002}} & \textcolor{biasgreen}{\textbf{0.979 ± 0.014}} \\
\bottomrule
\end{tabular}
}
\caption{Performance evaluation of fine-tuning strategies on UTKFace, FairFace, and WaterBird datasets, averaged over three random seeds. The best values are shown in \textbf{bold}. The scores that improve upon \textsc{Erm} by more than 0.005 are highlighted in \textcolor{biasgreen}{green}, while degradations greater than 0.005 are shown in \textcolor{biasred}{red}. Changes within ±0.005 of \textsc{Erm} score are considered insignificant.}
\label{tab:auc-performance-utk-fair-water}
\end{table*}

\begin{table*}[ht]
\centering
\resizebox{0.7\linewidth}{!}{
\begin{tabular}{llcccc}
\toprule
\textbf{Metric} & \textbf{Method} & \multicolumn{4}{c}{\textbf{CelebA}} \\
\cmidrule(lr){3-6}
& & Bald $\times$ Gender & Black hair $\times$ Gender & Eyeglasses $\times$ Gender & Smiling $\times$ Gender \\
\midrule
\multirow{5}{*}{\texttt{ROC} ($\uparrow$)}
& \textsc{Erm} & \cellcolor{gray!20}0.995 ± 0.000 & \cellcolor{gray!20}0.965 ± 0.000 & \cellcolor{gray!20}0.998 ± 0.000 & \cellcolor{gray!20}0.982 ± 0.000 \\
& \textsc{Unl} & 0.994 ± 0.001 & 0.963 ± 0.000 & 0.998 ± 0.000 & 0.981 ± 0.000 \\
& \textsc{Adv} & \textbf{0.997 ± 0.000} & 0.967 ± 0.000 & \textbf{0.999 ± 0.000} & 0.983 ± 0.000 \\
& \textsc{Dis} & \textbf{0.997 ± 0.000} & \textbf{0.968 ± 0.000} & \textbf{0.999 ± 0.000} & \textbf{0.984 ± 0.000} \\
& \textsc{Ent} & 0.990 ± 0.001 & \textbf{0.968 ± 0.000} & 0.998 ± 0.001 & 0.984 ± 0.001 \\
\midrule
\multirow{5}{*}{$\Delta$\texttt{ROC} ($\downarrow$)}
& \textsc{Erm} & \cellcolor{gray!20}0.008 ± 0.003 & \cellcolor{gray!20}0.016 ± 0.002 & \cellcolor{gray!20}0.003 ± 0.001 & \cellcolor{gray!20}0.008 ± 0.001 \\
& \textsc{Unl} & 0.008 ± 0.004 & 0.017 ± 0.001 & 0.002 ± 0.001 & 0.010 ± 0.001 \\
& \textsc{Adv} & \textcolor{biasred}{0.019 ± 0.012} & 0.015 ± 0.001 & \textbf{0.001 ± 0.000} & 0.008 ± 0.001 \\
& \textsc{Dis} & \textbf{0.005 ± 0.002} & 0.015 ± 0.001 & \textbf{0.001 ± 0.000} & \textbf{0.007 ± 0.001} \\
& \textsc{Ent} & 0.010 ± 0.002 & \textbf{0.014 ± 0.001} & 0.003 ± 0.001 & \textbf{0.007 ± 0.001} \\
\midrule
\multirow{5}{*}{\texttt{ROC} ratio ($\uparrow$)}
& \textsc{Erm} & \cellcolor{gray!20}0.992 ± 0.003 & \cellcolor{gray!20}0.983 ± 0.002 & \cellcolor{gray!20}0.997 ± 0.001 & \cellcolor{gray!20}0.991 ± 0.001 \\
& \textsc{Unl} & 0.992 ± 0.004 & 0.982 ± 0.001 & 0.998 ± 0.001 & 0.990 ± 0.001 \\
& \textsc{Adv} & \textcolor{biasred}{0.981 ± 0.012} & 0.984 ± 0.001 & \textbf{0.999 ± 0.000} & 0.992 ± 0.001 \\
& \textsc{Dis} & \textbf{0.995 ± 0.002} & 0.985 ± 0.001 & \textbf{0.999 ± 0.000} & \textbf{0.993 ± 0.001} \\
& \textsc{Ent} & 0.990 ± 0.002 & \textbf{0.986 ± 0.001} & 0.997 ± 0.001 & 0.992 ± 0.001 \\
\midrule
\midrule
\multirow{5}{*}{\texttt{PR} ($\uparrow$)}
& \textsc{Erm} & \cellcolor{gray!20}0.828 ± 0.010 & \cellcolor{gray!20}0.894 ± 0.001 & \cellcolor{gray!20}0.987 ± 0.001 & \cellcolor{gray!20}0.982 ± 0.000 \\
& \textsc{Unl} & \textcolor{biasred}{0.808 ± 0.005} & \textcolor{biasred}{0.888 ± 0.001} & 0.985 ± 0.001 & 0.982 ± 0.000 \\
& \textsc{Adv} & \textcolor{biasgreen}{0.861 ± 0.008} & \textcolor{biasgreen}{0.900 ± 0.001} & \textcolor{biasgreen}{\textbf{0.993 ± 0.001}} & 0.984 ± 0.000 \\
& \textsc{Dis} & \textcolor{biasgreen}{\textbf{0.869 ± 0.012}} & \textcolor{biasgreen}{\textbf{0.903 ± 0.001}} & \textcolor{biasgreen}{0.993 ± 0.002} & \textbf{0.985 ± 0.000} \\
& \textsc{Ent} & \textcolor{biasgreen}{0.851 ± 0.004} & \textcolor{biasgreen}{0.900 ± 0.001} & 0.988 ± 0.001 & 0.984 ± 0.001 \\
\midrule
\multirow{5}{*}{$\Delta$\texttt{PR} ($\downarrow$)}
& \textsc{Erm} & \cellcolor{gray!20}0.227 ± 0.109 & \cellcolor{gray!20}0.013 ± 0.003 & \cellcolor{gray!20}0.004 ± 0.003 & \cellcolor{gray!20}0.019 ± 0.000 \\
& \textsc{Unl} & 0.230 ± 0.077 & \textbf{0.009 ± 0.004} & 0.004 ± 0.002 & 0.021 ± 0.001 \\
& \textsc{Adv} & \textcolor{biasgreen}{\textbf{0.092 ± 0.047}} & 0.010 ± 0.002 & \textbf{0.004 ± 0.001} & 0.018 ± 0.000 \\
& \textsc{Dis} & \textcolor{biasgreen}{0.127 ± 0.013} & 0.011 ± 0.003 & 0.005 ± 0.002 & \textbf{0.017 ± 0.000} \\
& \textsc{Ent} & \textcolor{biasgreen}{0.136 ± 0.015} & 0.015 ± 0.002 & 0.009 ± 0.003 & 0.017 ± 0.001 \\
\midrule
\multirow{5}{*}{\texttt{PR} ratio ($\uparrow$)}
& \textsc{Erm} & \cellcolor{gray!20}0.743 ± 0.130 & \cellcolor{gray!20}0.986 ± 0.003 & \cellcolor{gray!20}0.996 ± 0.003 & \cellcolor{gray!20}0.981 ± 0.000 \\
& \textsc{Unl} & \textcolor{biasred}{0.718 ± 0.098} & \textbf{0.990 ± 0.004} & 0.996 ± 0.002 & 0.979 ± 0.001 \\
& \textsc{Adv} & \textcolor{biasgreen}{\textbf{0.902 ± 0.049}} & 0.989 ± 0.002 & \textbf{0.996 ± 0.001} & 0.982 ± 0.000 \\
& \textsc{Dis} & \textcolor{biasgreen}{0.868 ± 0.012} & 0.987 ± 0.003 & 0.995 ± 0.002 & \textbf{0.983 ± 0.000} \\
& \textsc{Ent} & \textcolor{biasgreen}{0.858 ± 0.010} & 0.983 ± 0.002 & 0.991 ± 0.003 & 0.983 ± 0.001 \\
\bottomrule
\end{tabular}
}
\caption{Performance evaluation of fine-tuning strategies on CelebA dataset, averaged over three random seeds. The best values are shown in \textbf{bold}. The scores that improve upon \textsc{Erm} by more than 0.005 are highlighted in \textcolor{biasgreen}{green}, while degradations greater than 0.005 are shown in \textcolor{biasred}{red}. Changes within ±0.005 of the \textsc{Erm} score are considered insignificant.}
\label{tab:auc-performance-celeba-1}
\end{table*}

\begin{table*}[ht]
\centering
\resizebox{0.7\linewidth}{!}{
\begin{tabular}{llcccc}
\toprule
\textbf{Metric} & \textbf{Method} & \multicolumn{4}{c}{\textbf{CelebA}} \\
\cmidrule(lr){3-6}
& & Wearing hat $\times$ Gender & Young $\times$ Gender & Attractive $\times$ Gender & Blond hair $\times$ Gender \\
\midrule
\multirow{5}{*}{\texttt{ROC} ($\uparrow$)}
& \textsc{Erm} & \cellcolor{gray!20}0.997 ± 0.000 & \cellcolor{gray!20}0.941 ± 0.001 & \cellcolor{gray!20}0.917 ± 0.001 & \cellcolor{gray!20}0.986 ± 0.000 \\
& \textsc{Unl} & 0.996 ± 0.000 & 0.940 ± 0.001 & 0.915 ± 0.000 & 0.986 ± 0.001 \\
& \textsc{Adv} & 0.998 ± 0.000 & 0.946 ± 0.001 & 0.919 ± 0.001 & 0.988 ± 0.000 \\
& \textsc{Dis} & \textbf{0.999 ± 0.000} & \textcolor{biasgreen}{\textbf{0.948 ± 0.001}} & \textbf{0.922 ± 0.001} & \textbf{0.989 ± 0.000} \\
& \textsc{Ent} & 0.996 ± 0.000 & \textcolor{biasgreen}{0.948 ± 0.002} & 0.921 ± 0.001 & 0.988 ± 0.000 \\
\midrule
\multirow{5}{*}{$\Delta$\texttt{ROC} ($\downarrow$)}
& \textsc{Erm} & \cellcolor{gray!20}0.002 ± 0.001 & \cellcolor{gray!20}0.006 ± 0.003 & \cellcolor{gray!20}0.004 ± 0.002 & \cellcolor{gray!20}0.005 ± 0.002 \\
& \textsc{Unl} & \textbf{0.000 ± 0.000} & \textbf{0.005 ± 0.001} & 0.005 ± 0.002 & 0.005 ± 0.002 \\
& \textsc{Adv} & 0.001 ± 0.001 & 0.006 ± 0.003 & 0.004 ± 0.002 & 0.002 ± 0.002 \\
& \textsc{Dis} & \textbf{0.000 ± 0.000} & 0.007 ± 0.003 & 0.004 ± 0.002 & \textbf{0.002 ± 0.001} \\
& \textsc{Ent} & 0.003 ± 0.001 & 0.006 ± 0.003 & \textbf{0.004 ± 0.001} & 0.005 ± 0.002 \\
\midrule
\multirow{5}{*}{\texttt{ROC} ratio ($\uparrow$)}
& \textsc{Erm} & \cellcolor{gray!20}0.998 ± 0.001 & \cellcolor{gray!20}0.994 ± 0.003 & \cellcolor{gray!20}0.995 ± 0.002 & \cellcolor{gray!20}0.994 ± 0.002 \\
& \textsc{Unl} & \textbf{1.000 ± 0.000} & \textbf{0.995 ± 0.001} & 0.994 ± 0.002 & 0.995 ± 0.003 \\
& \textsc{Adv} & 0.999 ± 0.001 & 0.994 ± 0.003 & 0.995 ± 0.002 & 0.998 ± 0.002 \\
& \textsc{Dis} & \textbf{1.000 ± 0.000} & 0.993 ± 0.003 & \textbf{0.996 ± 0.002} & \textbf{0.998 ± 0.001} \\
& \textsc{Ent} & 0.997 ± 0.001 & 0.994 ± 0.003 & 0.995 ± 0.001 & 0.995 ± 0.002 \\
\midrule
\midrule
\multirow{5}{*}{\texttt{PR} ($\uparrow$)}
& \textsc{Erm} & \cellcolor{gray!20}0.961 ± 0.002 & \cellcolor{gray!20}0.980 ± 0.001 & \cellcolor{gray!20}0.922 ± 0.001 & \cellcolor{gray!20}0.931 ± 0.002 \\
& \textsc{Unl} & 0.958 ± 0.001 & 0.980 ± 0.001 & 0.920 ± 0.001 & 0.928 ± 0.003 \\
& \textsc{Adv} & \textcolor{biasgreen}{0.971 ± 0.001} & 0.982 ± 0.001 & 0.924 ± 0.001 & 0.936 ± 0.002 \\
& \textsc{Dis} & \textcolor{biasgreen}{\textbf{0.975 ± 0.001}} & 0.983 ± 0.001 & \textbf{0.927 ± 0.001} & \textcolor{biasgreen}{\textbf{0.939 ± 0.001}} \\
& \textsc{Ent} & \textcolor{biasred}{0.942 ± 0.005} & \textbf{0.983 ± 0.000} & 0.926 ± 0.000 & 0.931 ± 0.002 \\
\midrule
\multirow{5}{*}{$\Delta$\texttt{PR} ($\downarrow$)}
& \textsc{Erm} & \cellcolor{gray!20}0.037 ± 0.003 & \cellcolor{gray!20}0.034 ± 0.001 & \cellcolor{gray!20}0.170 ± 0.003 & \cellcolor{gray!20}0.356 ± 0.016 \\
& \textsc{Unl} & 0.038 ± 0.007 & 0.035 ± 0.000 & 0.165 ± 0.004 & \textcolor{biasred}{0.366 ± 0.014} \\
& \textsc{Adv} & \textcolor{biasgreen}{0.026 ± 0.005} & 0.031 ± 0.000 & \textcolor{biasgreen}{0.163 ± 0.003} & \textcolor{biasgreen}{0.330 ± 0.011} \\
& \textsc{Dis} & \textcolor{biasgreen}{\textbf{0.023 ± 0.003}} & 0.031 ± 0.001 & \textcolor{biasgreen}{0.160 ± 0.003} & \textcolor{biasgreen}{0.314 ± 0.013} \\
& \textsc{Ent} & \textcolor{biasred}{0.043 ± 0.010} & \textbf{0.029 ± 0.000} & \textcolor{biasgreen}{\textbf{0.158 ± 0.003}} & \textcolor{biasgreen}{\textbf{0.294 ± 0.021}} \\
\midrule
\multirow{5}{*}{\texttt{PR} ratio ($\uparrow$)}
& \textsc{Erm} & \cellcolor{gray!20}0.962 ± 0.003 & \cellcolor{gray!20}0.965 ± 0.001 & \cellcolor{gray!20}0.821 ± 0.003 & \cellcolor{gray!20}0.623 ± 0.017 \\
& \textsc{Unl} & 0.961 ± 0.007 & 0.964 ± 0.000 & 0.825 ± 0.004 & \textcolor{biasred}{0.611 ± 0.015} \\
& \textsc{Adv} & \textcolor{biasgreen}{0.973 ± 0.005} & 0.969 ± 0.000 & \textcolor{biasgreen}{0.828 ± 0.003} & \textcolor{biasgreen}{0.651 ± 0.011} \\
& \textsc{Dis} & \textcolor{biasgreen}{\textbf{0.976 ± 0.003}} & 0.969 ± 0.001 & \textcolor{biasgreen}{0.832 ± 0.003} & \textcolor{biasgreen}{0.670 ± 0.013} \\
& \textsc{Ent} & \textcolor{biasred}{0.955 ± 0.011} & \textbf{0.970 ± 0.000} & \textcolor{biasgreen}{\textbf{0.834 ± 0.003}} & \textcolor{biasgreen}{\textbf{0.687 ± 0.022}} \\
\bottomrule
\end{tabular}
}
\caption{Performance evaluation of fine-tuning strategies on CelebA dataset, averaged over three random seeds. The best values are shown in \textbf{bold}. The scores that improve upon \textsc{Erm} by more than 0.005 are highlighted in \textcolor{biasgreen}{green}, while degradations greater than 0.005 are shown in \textcolor{biasred}{red}. Changes within ±0.005 of the \textsc{Erm} score are considered insignificant.}
\label{tab:auc-performance-celeba-2}
\end{table*}

\textbf{Orthogonality-based disentanglement (\textsc{Dis}).} This method aims to prevent sensitive attribute information from being entangled with downstream task representations by explicitly enforcing orthogonality between them. We first train a sensitive adapter $\theta^{(\text{sen})}_{\textsc{Erm}}$ and its classification head $w^{(\text{sen})}_{\textsc{Erm}}$ using Eq.~\eqref{eq:erm-sen-obj}. Keeping both $\theta^{(\text{pre})}$ and $\theta^{(\text{sen})}_{\textsc{Erm}}$ frozen, we then train a downstream task adapter $\theta^{(\text{task})}_{\textsc{Dis}}$ and its classification head $w^{(\text{task})}_{\textsc{Dis}}$ on $\mathcal{D}^{(\text{task})}$ using a composite objective that combines the standard cross-entropy loss, and an orthogonality regularizer that penalizes cosine similarity between task and sensitive representations. For an input $x$, let $h_{\text{task}}(x) = f(x; \theta^{(\text{pre})} \oplus \theta^{(\text{task})})$ and $h_{\text{sen}}(x) = f(x; \theta^{(\text{pre})} \oplus \theta^{(\text{sen})}_{\textsc{Erm}})$ denote the task and sensitive representations, respectively, where $f(\cdot)$ is the frozen backbone plus adapter up to the representation layer. The orthogonality regularizer is defined as $R_{\textsc{Dis}} (\theta^{(\text{pre})} \oplus \theta^{(\text{task})}, \theta^{(\text{pre})} \oplus \theta^{(\text{sen})}_{\textsc{Erm}}; \mathcal{D}^{(\text{task})}) = \frac{1}{|\mathcal{D}^{(\text{task})}|} \sum_{x \in \mathcal{D}^{(\text{task})}} \cos (h_{\text{task}}(x), h_{\text{sen}}(x))$, where $\cos (u,v) = \frac{u^\top v}{\norm{u} \cdot \norm{v}}$ is the cosine similarity. The complete training objective is
\begin{align}
& (\theta^{(\text{task})}_{\textsc{Dis}}, w^{(\text{task})}_{\textsc{Dis}}) \gets \argmin_{\theta^{(\text{task})}, w^{(\text{task})}} \big[ \lambda_{\textsc{Norm}} \cdot R_{\textsc{Norm}} (\theta^{(\text{task})}) \nonumber \\
& \quad \quad + \lambda_{\textsc{Dis}} \cdot R_{\textsc{Dis}} (\theta^{(\text{pre})} \oplus \theta^{(\text{task})}, \theta^{(\text{pre})} \oplus \theta^{(\text{sen})}_{\textsc{Erm}}; \mathcal{D}^{(\text{task})}) \nonumber \\
& \quad \quad + \ell_{\text{CE}} (\theta^{(\text{pre})} \oplus \theta^{(\text{task})}, w^{(\text{task})}; \mathcal{D}^{(\text{task})}) \big], \label{eq:orth-obj}
\end{align}
where $\lambda_{\textsc{Dis}}$ controls the strength of the orthogonality constraint, encouraging the downstream adapter to capture information complementary to (and thus disentangled from) the sensitive attribute.

\textbf{Entropy maximization regularization (\textsc{Ent}).} This method follows the first two stages of \textsc{Adv} but replaces adversarial training with entropy maximization to encourage uncertainty in sensitive attribute predictions. After pre-training $(\theta^{(\text{task})}_{\textsc{Erm}}, w^{(\text{task})}_{\textsc{Erm}})$ on $\mathcal{D}^{(\text{task})}$ and training $w^{(\text{task})}_{\textsc{sen}}$ on $\mathcal{D}^{(\text{sen})}$, we fine-tune $(\theta^{(\text{task})}, w^{(\text{task})})$---initialized from stage one---using $R_\textsc{Ent} (\theta^{(\text{pre})} \oplus \theta^{(\text{task})}, w^{(\text{sen})}_{\textsc{Erm}}; \mathcal{D}^{(\text{task})}) = - \frac{1}{|\mathcal{D}^{(\text{task})}|} \sum_{x \in \mathcal{D}^{(\text{task})}} H(p_\text{sen}(x))$, where $p_\text{sen}(x) = \mathrm{softmax}(w^{(\text{sen})}_{\textsc{Erm}} \circ h_\text{task}(x))$ is the sensitive attribute probability distribution predicted by the frozen sensitive head, $h_\text{task}(x) = f(x; \theta^{(\text{pre})} \oplus \theta^{(\text{task})})$ is the task representation, and $H(p) = -\sum_{c} p_c \log p_c$ is the Shannon entropy. Maximizing entropy pushes sensitive predictions toward uniformity, reducing leakage. The final training objective is 
\begin{align}
& (\theta^{(\text{task})}_\textsc{Ent}, w^{(\text{task})}_\textsc{Ent}) \gets \argmin_{\theta^{(\text{task})}, w^{(\text{task})}} \big[ \lambda_{\textsc{Norm}} \cdot R_{\textsc{Norm}} (\theta^{(\text{task})}) \nonumber \\
& \quad \quad + \lambda_{\textsc{Ent}} \cdot R_\textsc{Ent} (\theta^{(\text{pre})} \oplus \theta^{(\text{task})}, w^{(\text{sen})}_{\textsc{Erm}}; \mathcal{D}^{(\text{task})}) \nonumber \\
& \quad \quad + \ell_{\text{CE}} (\theta^{(\text{pre})} \oplus \theta^{(\text{task})}, w^{(\text{task})}; \mathcal{D}^{(\text{task})}) \big], \label{eq:ent-step-3}
\end{align}
where $(\theta^{(\text{task})}, w^{(\text{task})})$ are initialized from stage one, and $\theta^{(\text{pre})}$ and $w^{(\text{task})}_{\textsc{Erm}}$ remain frozen.


\section{Experiments}
\label{sec:experiments}

In this section, we conduct a comprehensive evaluation of the fine-tuning strategies introduced in Section~\ref{sec:bias-mitigation}, assessing both utility and fairness performance.  

\subsection{Experimental Setup}

\textbf{Datasets and Base Model.} We evaluate on widely used image datasets in the algorithmic fairness literature: CelebA~\cite{liu2015deep}, UTK-Face~\cite{zhang2017age}, FairFace~\cite{karkkainen2021fairface}, and WaterBirds~\cite{sagawadistributionally}. CelebA comprises 202{,}599 face images with 40 binary attribute labels. UTK-Face contains 20{,}000 face images annotated with gender, age, and race. FairFace contains 108{,}501 face images with gender, age, and race annotations and features a race-balanced distribution. WaterBird contains 11{,}708 bird images annotated with bird type and background. The downstream and sensitive attributes of each dataset are summarized in Table~4 in the appendix of the supplementary material. All images are resized to $224 \times 224$ to match our base model's input and are normalized prior to training. We randomly split each dataset into training (70\%), validation (15\%), and test (15\%) sets. We use an ImageNet-pretrained Vision Transformer (ViT-Base)~\cite{dosovitskiy2021an}, specifically the \texttt{vit-base-patch16-224-in21k} version with 86.6M parameters\footnote{\url{https://huggingface.co/google/vit-base-patch16-224-in21k}}. 

\textbf{Training Details.} We employ a class-balanced sampling strategy and optimize the models using the AdamW optimizer~\cite{loshchilovdecoupled} with a batch size of 256, a learning rate of 1e-4, and a weight decay of 5e-4. Training runs for 10 epochs with a CosineAnnealingLR scheduler, where $T_{\text{max}} = 5$. We enable early stopping with a patience of 5 consecutive epochs if validation accuracy remains within a 2\% band. For the LoRA adapters, we set the rank to 4, the alpha parameter to 8, and apply a dropout rate of 0.1. Our implementation utilizes the PEFT library from Hugging Face~\cite{peft}. Method-specific hyperparameters are fixed across datasets as follows: $\lambda_{\textsc{Norm}}=1\mathrm{e}{-5}$, $\lambda_{\textsc{Unl}}=1.0$, $\lambda_{\textsc{Adv}}=0.1$, $\lambda_{\textsc{Dis}}=0.1$, and $\lambda_{\textsc{Ent}}=1.0$. 

\subsection{Results}

We evaluate the fine-tuning strategies from Section~\ref{sec:bias-mitigation} using both utility and fairness metrics, implemented with \texttt{scikit-learn}~\cite{pedregosa2011scikit} and \texttt{Fairlearn}~\cite{weerts2023fairlearn}. Precision, recall, and false-positive rate (FPR) are macro-averaged across classes to accommodate both binary and multiclass settings. All results are averaged over three random seeds.

Tables~\ref{tab:auc-performance-utk-fair-water}, \ref{tab:auc-performance-celeba-1}, and \ref{tab:auc-performance-celeba-2} report AUC-based results, covering both utility (overall AUC) and fairness (subgroup-wise AUC ratio and difference). For AUROC, the \textsc{Erm} baseline exhibits strong utility across datasets (overall AUROC $>0.90$) with small subgroup disparities (difference $<0.05$; ratio $>0.95$). Averaging method-wise AUROC scores and their rankings across datasets (Figure~\ref{fig:auroc-average}), we observe the ordering $\textsc{Dis} \succ \textsc{Ent} \succ \textsc{Adv} \succ \textsc{Erm} \succ \textsc{Unl}$ for both overall AUROC and subgroup-wise AUROC ratio. Thus, aside from \textsc{Unl}, all fairness-aware fine-tuning methods improve upon \textsc{Erm}. Notably, \textsc{Unl}---effective in generative detoxification--does not translate to gains for classification.

For AUPRC, \textsc{Erm} remains generally strong but shows utility dips for UTKFace (Age), FairFace (Age), and CelebA (Bald, Black Hair), alongside larger subgroup disparities for UTKFace (Age), FairFace (Age), WaterBird (Type), and CelebA (Bald, Attractive, Blond Hair). In contrast to AUROC, \textsc{Dis}, \textsc{Ent}, and \textsc{Adv} yield more pronounced gains over \textsc{Erm} in both utility and fairness. Aggregating AUPRC scores and rankings across datasets (Figure~\ref{fig:auprc-average}) gives $\textsc{Dis} \succ \textsc{Ent} \succ \textsc{Adv} \succ \textsc{Erm} \succ \textsc{Unl}$ for overall AUPRC (utility) and $\textsc{Ent} \succ \textsc{Dis} \succ \textsc{Adv} \succ \textsc{Erm} \succ \textsc{Unl}$ for the subgroup-wise AUPRC ratio (fairness), again showing all fairness-aware methods outperform \textsc{Erm} except \textsc{Unl}.

Additional metrics appear in the appendix due to space constraints (Tables~5, 6, and 7), reporting overall utility and subgroup-wise ratio/difference for accuracy, precision, recall, and FPR. For accuracy, \textsc{Erm} is typically strong in utility but drops below $0.90$ on UTKFace (Age), FairFace (Age), and CelebA (Black Hair, Young, Attractive), with larger disparities (difference $>0.05$, ratio $<0.95$) for UTKFace (Age, Gender) and FairFace (Age, Gender). For precision, similar utility dips occur on UTKFace (Age), FairFace (Age), and CelebA (Bald, Black Hair, Wearing Hat, Young, Attractive, Blond Hair), with larger disparities for UTKFace (Age, Gender), FairFace (Age, Gender), and CelebA (Bald, Wearing Hat, Blond Hair). For recall, utility drops are observed on UTKFace (Age), FairFace (Age), and CelebA (Young, Attractive), with larger disparities for UTKFace (Age, Gender), FairFace (Age, Gender), and CelebA (Bald, Blond Hair). Across these metrics, \textsc{Dis} and \textsc{Ent} generally improve upon or match \textsc{Erm} in both utility and fairness, while \textsc{Adv} is competitive but less consistent.

FPR reveals an interesting pattern. In terms of overall FPR (utility), \textsc{Erm} is mostly solid but shows higher error (FPR $>0.10$) for CelebA (Young, Attractive), and larger subgroup differences (difference $>0.05$) for UTKFace (Gender), FairFace (Gender), and CelebA (Bald, Blond Hair). However, for the subgroup-wise FPR ratio, substantial bias (ratio $<0.95$) appears broadly across datasets and target–sensitive pairs. Moreover, fairness-aware methods often underperform \textsc{Erm} on this particular ratio metric, suggesting a tension among fairness objectives: improving one criterion can degrade another. Since we keep method-specific hyperparameters fixed across datasets, targeted tuning may help balance these trade-offs.

Overall, \textsc{Dis} and \textsc{Ent} maintain high downstream performance while substantially reducing bias where \textsc{Erm} shows disparities; \textsc{Adv} is competitive but less stable. In contrast, \textsc{Unl} offers limited benefits over \textsc{Erm} in classification, despite prior success in generative detoxification.

\section{Conclusion}
\label{sec:conclusion}

This work presents four fairness-aware fine-tuning methods designed for the practically relevant scenario where downstream task and sensitive attribute datasets are available separately. Through comprehensive experiments on standard fairness datasets, we demonstrate that while adversarial debiasing yields inconsistent improvements and sensitive unlearning proves ineffective for classification tasks, orthogonality-based disentanglement and entropy maximization approaches consistently reduce bias and often enhance overall utility compared to standard fine-tuning. However, our findings also reveal fundamental incompatibilities between fairness metrics, underscoring the inherent challenges of simultaneously optimizing multiple fairness objectives in parameter-efficient adaptation.

These promising findings open several avenues for future investigation. First, extending fairness-aware fine-tuning to intersectional fairness settings~\cite{gohar2023survey} by incorporating separate classification heads and sensitive adapters for each attribute in the intersection could address more complex bias patterns. Second, investigating the impact of distribution shift between task and sensitive attribute datasets, and developing robust fairness-aware methods that maintain performance under such shifts~\cite{shaosupervised}, represents a critical practical consideration. Finally, studying fairness-aware fine-tuning under privacy-constrained scenarios by incorporating differentially private methods for sharing LoRA adapters~\cite{yudifferentially,sunimproving} would enable broader deployment in sensitive applications.

\bibliography{main}

@String(PAMI = {IEEE Trans. Pattern Anal. Mach. Intell.})

@String(CVPR= {IEEE Conf. Comput. Vis. Pattern Recog.})

@String(ICCV= {Int. Conf. Comput. Vis.})

@String(NIPS= {Adv. Neural Inform. Process. Syst.})

@String(ICLR = {Int. Conf. Learn. Represent.})

@String(IJCAI = {IJCAI})

@String(AAAI = {AAAI})

@String(PAMI  = {IEEE TPAMI})

@String(CVPR  = {CVPR})

@String(ICCV  = {ICCV})

@String(NIPS  = {NeurIPS})

@String(ICLR  = {ICLR})

@String(ICML  = {ICML})

@String(AIES  = {AIES})

@String(KDD  = {KDD})

@String(ICDM  = {ICDM})

@String(JMLR  = {JMLR})

@String(EAAMO = {EAAMO})

@inproceedings{adel2019one,
  title={{One-Network Adversarial Fairness}},
  author={Adel, Tameem and Valera, Isabel and Ghahramani, Zoubin and Weller, Adrian},
  booktitle=AAAI,
  year={2019}
}

@inproceedings{agarwal2018reductions,
  title={{A Reductions Approach to Fair Classification}},
  author={Agarwal, Alekh and Beygelzimer, Alina and Dud{\'\i}k, Miroslav and Langford, John and Wallach, Hanna},
  booktitle=ICML,
  year={2018}
}

@inproceedings{agarwal2019fair,
  title={{Fair Regression: Quantitative Definitions and Reduction-based Algorithms}},
  author={Agarwal, Alekh and Dud{\'\i}k, Miroslav and Wu, Zhiwei Steven},
  booktitle=ICML,
  year={2019}
}

@inproceedings{ali2023evaluating,
  title={{Evaluating the Fairness of Discriminative Foundation Models in Computer Vision}},
  author={Ali, Junaid and Kleindessner, Matth{\"a}us and Wenzel, Florian and Budhathoki, Kailash and Cevher, Volkan and Russell, Chris},
  booktitle=AIES,
  year={2023}
}

@article{awais2025foundation,
  title={{Foundation Models Defining a New Era in Vision: a Survey and Outlook}},
  author={Awais, Muhammad and others},
  journal=PAMI,
  year={2025}
}

@book{barocas2023fairness,
  title={{Fairness and Machine Learning: Limitations and Opportunities}},
  author={Barocas, Solon and Hardt, Moritz and Narayanan, Arvind},
  year={2023},
  publisher={MIT press}
}

@article{bellamy2019ai,
  title={{AI Fairness 360: An Extensible Toolkit for Detecting and Mitigating Algorithmic Bias}},
  author={Bellamy, Rachel KE and others},
  journal={IBM Journal of Research and Development},
  year={2019}
}

@article{bommasani2021opportunities,
  title={{On the Opportunities and Risks of Foundation Models}},
  author={Bommasani, Rishi and others},
  journal={arXiv preprint arXiv:2108.07258},
  year={2021}
}

@article{calmon2017optimized,
  title={{Optimized Pre-Processing for Discrimination Prevention}},
  author={Calmon, Flavio and Wei, Dennis and Vinzamuri, Bhanukiran and Natesan Ramamurthy, Karthikeyan and Varshney, Kush R},
  journal=NIPS,
  year={2017}
}

@article{locatello2019fairness,
  title={{On the Fairness of Disentangled Representations}},
  author={Locatello, Francesco and Abbati, Gabriele and Rainforth, Thomas and Bauer, Stefan and Sch{\"o}lkopf, Bernhard and Bachem, Olivier},
  journal=NIPS,
  year={2019}
}

@article{caton2020fairness,
author = {Caton, Simon and Haas, Christian},
title = {{Fairness in Machine Learning: A Survey}},
year = {2024},
journal = {ACM Comput. Surv.},
}

@article{chouldechova2017fair,
  title={{Fair Prediction with Disparate Impact: A Study of Bias in Recidivism Prediction Instruments}},
  author={Chouldechova, Alexandra},
  journal={Big data},
  year={2017}
}

@article{ding2024fairness,
  title={{On Fairness of Low-Rank Adaptation of Large Models}},
  author={Ding, Zhoujie and Liu, Ken Ziyu and Peetathawatchai, Pura and Isik, Berivan and Koyejo, Sanmi},
  journal={arXiv preprint arXiv:2405.17512},
  year={2024}
}

@inproceedings{
dosovitskiy2021an,
title={{An Image is Worth 16x16 Words: Transformers for Image Recognition at Scale}},
author={Alexey Dosovitskiy and others},
booktitle=ICLR,
year={2021}
}

@inproceedings{duttfairtune,
  title={{FairTune: Optimizing Parameter Efficient Fine Tuning for Fairness in Medical Image Analysis}},
  author={Dutt, Raman and Bohdal, Ondrej and Tsaftaris, Sotirios A and Hospedales, Timothy},
  booktitle=ICLR,
  year={2024}
}

@inproceedings{dwork2012fairness,
  title={{Fairness Through Awareness}},
  author={Dwork, Cynthia and Hardt, Moritz and Pitassi, Toniann and Reingold, Omer and Zemel, Richard},
  booktitle={ITCS},
  year={2012}
}

@inproceedings{feldman2015certifying,
  title={{Certifying and Removing Disparate Impact}},
  author={Feldman, Michael and Friedler, Sorelle A and Moeller, John and Scheidegger, Carlos and Venkatasubramanian, Suresh},
  booktitle=KDD,
  year={2015}
}

@article{han2024parameter,
  title={{Parameter-Efficient Fine-Tuning for Large Models: A Comprehensive Survey}},
  author={Han, Zeyu and Gao, Chao and Liu, Jinyang and Zhang, Jeff and Zhang, Sai Qian},
  journal={arXiv preprint arXiv:2403.14608},
  year={2024}
}

@article{hardt2016equality,
  title={{Equality of Opportunity in Supervised Learning}},
  author={Hardt, Moritz and Price, Eric and Srebro, Nati},
  journal=NIPS,
  year={2016}
}

@article{hort2024bias,
  title={{Bias Mitigation for Machine Learning Classifiers: A Comprehensive Survey}},
  author={Hort, Max and Chen, Zhenpeng and Zhang, Jie M and Harman, Mark and Sarro, Federica},
  journal={ACM Journal on Responsible Computing},
  year={2024}
}

@inproceedings{hu2022lora,
  title={{LoRA: Low-Rank Adaptation of Large Language Models}},
  author={Hu, Edward J and Wallis, Phillip and Allen-Zhu, Zeyuan and Li, Yuanzhi and Wang, Shean and Wang, Lu and Chen, Weizhu and others},
  booktitle=ICLR,
  year={2022}
}

@article{kamiran2012data,
  title={{Data Pre-Processing Techniques for Classification without Discrimination}},
  author={Kamiran, Faisal and Calders, Toon},
  journal={Knowledge and Information Systems},
  year={2012}
}

@inproceedings{kamiran2012decision,
  title={{Decision Theory for Discrimination-Aware Classification}},
  author={Kamiran, Faisal and Karim, Asim and Zhang, Xiangliang},
  booktitle=ICDM,
  year={2012}
}

@inproceedings{kamishima2012fairness,
  title={{Fairness-Aware Classifier with Prejudice Remover Regularizer}},
  author={Kamishima, Toshihiro and Akaho, Shotaro and Asoh, Hideki and Sakuma, Jun},
  booktitle={ECML PKDD},
  year={2012}
}

@inproceedings{liu2015deep,
  title={{Deep Learning Face Attributes in the Wild}},
  author={Liu, Ziwei and Luo, Ping and Wang, Xiaogang and Tang, Xiaoou},
  booktitle=ICCV,
  year={2015}
}

@inproceedings{karkkainen2021fairface,
  title={{FairFace: Face Attribute Dataset for Balanced Race, Gender, and Age for Bias Measurement and Mitigation}},
  author={Karkkainen, Kimmo and Joo, Jungseock},
  booktitle={WACV},
  year={2021}
}

@inproceedings{sagawadistributionally,
  title={{Distributionally Robust Neural Networks for Group Shifts: On the Importance of Regularization for Worst-Case Generalization}},
  author={Sagawa, Shiori and Koh, Pang Wei and Hashimoto, Tatsunori B and Liang, Percy},
  booktitle=ICLR,
  year={2020}
}

@article{mao2023last,
  title={{Last-Layer Fairness Fine-tuning is Simple and Effective for Neural Networks}},
  author={Mao, Yuzhen and Deng, Zhun and Yao, Huaxiu and Ye, Ting and Kawaguchi, Kenji and Zou, James},
  journal={arXiv preprint arXiv:2304.03935},
  year={2023}
}

@article{mehrabi2021survey,
  title={{A Survey on Bias and Fairness in Machine Learning}},
  author={Mehrabi, Ninareh and Morstatter, Fred and Saxena, Nripsuta and Lerman, Kristina and Galstyan, Aram},
  journal={ACM computing surveys (CSUR)},
  year={2021}
}

@article{pleiss2017fairness,
  title={{On Fairness and Calibration}},
  author={Pleiss, Geoff and Raghavan, Manish and Wu, Felix and Kleinberg, Jon and Weinberger, Kilian Q},
  journal=NIPS,
  year={2017}
}

@article{sukumaran2024fairlora,
  title={{FairLoRA: Unpacking Bias Mitigation in Vision Models with Fairness-Driven Low-Rank Adaptation}},
  author={Sukumaran, Rohan and Feizi, Aarash and Romero-Sorian, Adriana and Farnadi, Golnoosh},
  journal={arXiv preprint arXiv:2410.17358},
  year={2024}
}

@inproceedings{verma2018fairness,
  title={{Fairness Definitions Explained}},
  author={Verma, Sahil and Rubin, Julia},
  booktitle={Proceedings of the International Workshop on Software Fairness},
  year={2018}
}

@article{wan2023processing,
  title={{In-Processing Modeling Techniques for Machine Learning Fairness: A Survey}},
  author={Wan, Mingyang and Zha, Daochen and Liu, Ninghao and Zou, Na},
  journal={ACM Transactions on Knowledge Discovery from Data},
  year={2023}
}

@article{weerts2023fairlearn,
  title={{Fairlearn: Assessing and Improving Fairness of AI Systems}},
  author={Weerts, Hilde and Dud{\'\i}k, Miroslav and Edgar, Richard and Jalali, Adrin and Lutz, Roman and Madaio, Michael},
  journal=JMLR,
  year={2023}
}

@article{pedregosa2011scikit,
  title={{Scikit-learn: Machine learning in Python}},
  author={Pedregosa, Fabian and others},
  journal=JMLR,
  year={2011}
}

@article{ganin2016domain,
  title={{Domain-Adversarial Training of Neural Networks}},
  author={Ganin, Yaroslav and Ustinova, Evgeniya and Ajakan, Hana and Germain, Pascal and Larochelle, Hugo and Laviolette, Fran{\c{c}}ois and March, Mario and Lempitsky, Victor},
  journal=JMLR,
  year={2016}
}

@inproceedings{zhang2018mitigating,
  title={{Mitigating Unwanted Biases with Adversarial Learning}},
  author={Zhang, Brian Hu and Lemoine, Blake and Mitchell, Margaret},
  booktitle=AIES,
  year={2018}
}

@article{zhang2023composing,
  title={{Composing Parameter-Efficient Modules with Arithmetic Operations}},
  author={Zhang, Jinghan and Chen, Shiqi and Liu, Junteng and He, Junxian},
  journal=NIPS,
  year={2023}
}

@inproceedings{zhang2023adalora,
  title={{Adaptive Budget Allocation for Parameter-Efficient Fine-Tuning}},
  author={Zhang, Qingru and Chen, Minshuo and Bukharin, Alexander and He, Pengcheng and Cheng, Yu and Chen, Weizhu and Zhao, Tuo},
  booktitle=ICLR,
  year={2023}
}

@inproceedings{zhang2017age,
  title={{Age Progression/Regression by Conditional Adversarial Autoencoder}},
  author={Zhang, Zhifei and Song, Yang and Qi, Hairong},
  booktitle=CVPR,
  year={2017}
}

@article{zhao2023survey,
  title={{A Survey of Large Language Models}},
  author={Zhao, Wayne Xin and others},
  journal={arXiv preprint arXiv:2303.18223},
  year={2023}
}

@article{zhou2024comprehensive,
  title={{A Comprehensive Survey on Pretrained Foundation Models: A History from BERT to ChatGPT}},
  author={Zhou, Ce and others},
  journal={International Journal of Machine Learning and Cybernetics},
  year={2024}
}

@inproceedings{creager2019flexibly,
  title={{Flexibly Fair Representation Learning by Disentanglement}},
  author={Creager, Elliot and Madras, David and Jacobsen, J{\"o}rn-Henrik and Weis, Marissa and Swersky, Kevin and Pitassi, Toniann and Zemel, Richard},
  booktitle=ICML,
  year={2019}
}

@inproceedings{yudifferentially,
  title={{Differentially Private Fine-tuning of Language Models}},
  author={Yu, Da and others},
  booktitle=ICLR,
  year={2022}
}

@inproceedings{sunimproving,
  title={{Improving LoRA in Privacy-Preserving Federated Learning}},
  author={Sun, Youbang and Li, Zitao and Li, Yaliang and Ding, Bolin},
  booktitle=ICLR,
  year={2024}
}

@inproceedings{loshchilovdecoupled,
  title={{Decoupled Weight Decay Regularization}},
  author={Loshchilov, Ilya and Hutter, Frank},
  booktitle=ICLR,
  year={2024}
}

@Misc{peft,
  title={{PEFT: State-of-the-art Parameter-Efficient Fine-Tuning methods}},
  author={Sourab Mangrulkar and Sylvain Gugger and Lysandre Debut and Younes Belkada and Sayak Paul and Benjamin Bossan},
  xhowpublished={\url{https://github.com/huggingface/peft}},
  year={2022}
}

@inproceedings{ashurst2023fairness,
  title={{Fairness Without Demographic Data: A Survey of Approaches}},
  author={Ashurst, Carolyn and Weller, Adrian},
  booktitle=EAAMO,
  xpages={1--12},
  year={2023}
}

@inproceedings{kim2020fact,
  title={{FACT: A Diagnostic for Group Fairness Trade-offs}},
  author={Kim, Joon Sik and Chen, Jiahao and Talwalkar, Ameet},
  booktitle=ICML,
  xpages={5264--5274},
  year={2020},
  xorganization={PMLR}
}

@article{berk2021fairness,
  title={{Fairness in Criminal Justice Risk Assessments: The State of the Art}},
  author={Berk, Richard and Heidari, Hoda and Jabbari, Shahin and Kearns, Michael and Roth, Aaron},
  journal={Sociological Methods \& Research},
  xvolume={50},
  xnumber={1},
  xpages={3--44},
  year={2021},
  xpublisher={Sage Publications Sage CA: Los Angeles, CA}
}

@article{zhang2025lori,
  title={{LoRI: Reducing Cross-Task Interference in Multi-Task LowRank Adaptation}},
  author={Zhang, Juzheng and You, Jiacheng and Panda, Ashwinee and Goldstein, Tom},
  journal={arXiv preprint arXiv:2504.07448},
  year={2025}
}

@inproceedings{gohar2023survey,
  title={{A Survey on Intersectional Fairness in Machine Learning: Notions, Mitigation, and Challenges}},
  author={Gohar, Usman and Cheng, Lu},
  booktitle=IJCAI,
  year={2023}
}

@inproceedings{shaosupervised,
  title={{Supervised Algorithmic Fairness in Distribution Shifts: A Survey}},
  author={Shao, Minglai and Li, Dong and Zhao, Chen and Wu, Xintao and Lin, Yujie and Tian, Qin},
  booktitle=IJCAI,
  year={2024}
}

@inproceedings{roy2019mitigating,
  title={{Mitigating Information Leakage in Image Representations: A Maximum Entropy Approach}},
  author={Roy, Proteek Chandan and Boddeti, Vishnu Naresh},
  booktitle=CVPR,
  year={2019}
}
\bibliographystyle{icml2026}

\newpage
\appendix
\onecolumn

\clearpage
\appendix

\section{Additional Details and Results} 
\label{app:results-details}

\begin{figure*}[ht]
    \centering
    \begin{subfigure}[b]{0.24\linewidth}
        \centering
        \includegraphics[width=0.9\linewidth]{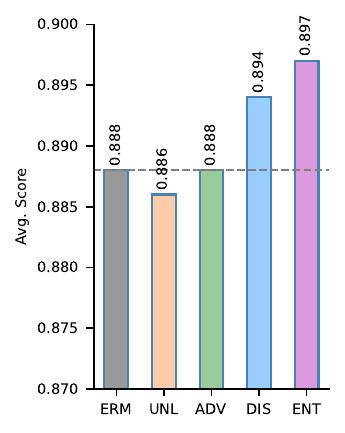}
        \caption{Overall Acc Score ($\uparrow$)}
        \label{fig:avg-acc}
    \end{subfigure}
    \begin{subfigure}[b]{0.24\linewidth}
        \centering
        \includegraphics[width=0.9\linewidth]{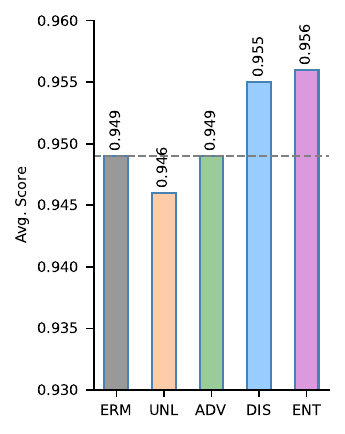}
        \caption{Acc Ratio Score ($\uparrow$)}
        \label{fig:avg-acc-ratio}
    \end{subfigure}
    \begin{subfigure}[b]{0.24\linewidth}
        \centering
        \includegraphics[width=0.9\linewidth]{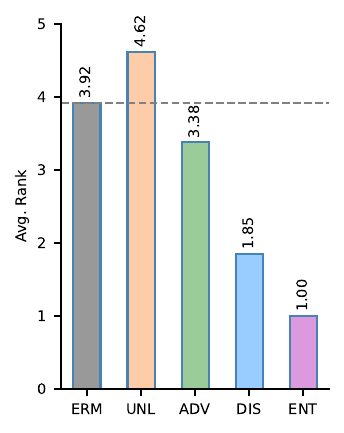}
        \caption{Overall Acc Rank ($\downarrow$)}
        \label{fig:avg-acc-rank}
    \end{subfigure}
    \begin{subfigure}[b]{0.24\linewidth}
        \centering
        \includegraphics[width=0.9\linewidth]{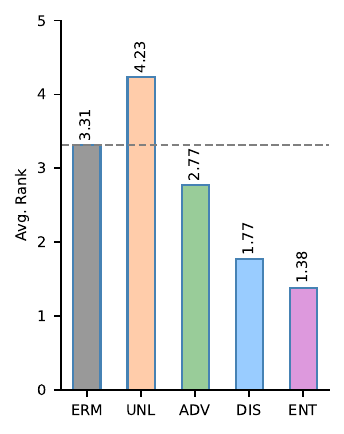}
        \caption{Acc Ratio Rank ($\downarrow$)}
        \label{fig:avg-acc-ratio-rank}
    \end{subfigure}
    \caption{Average overall accuracy and group-wise accuracy ratio (ratio of the best- to worst-performing subgroup) scores and ranks for each method, aggregated over all experiments.}
    \label{fig:acc-average}
\end{figure*}

\begin{figure*}[ht]
    \centering
    \begin{subfigure}[b]{0.24\linewidth}
        \centering
        \includegraphics[width=0.9\linewidth]{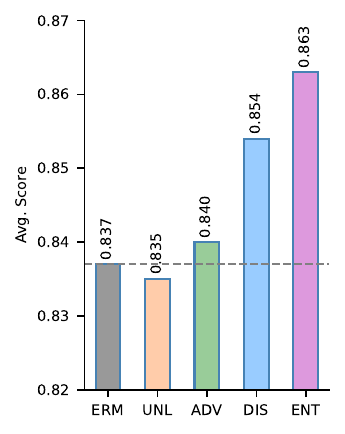}
        \caption{Overall PPV Score ($\uparrow$)}
        \label{fig:avg-ppv}
    \end{subfigure}
    \begin{subfigure}[b]{0.24\linewidth}
        \centering
        \includegraphics[width=0.9\linewidth]{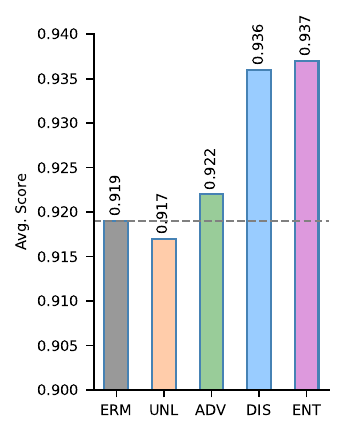}
        \caption{PPV Ratio Score ($\uparrow$)}
        \label{fig:avg-ppv-ratio}
    \end{subfigure}
    \begin{subfigure}[b]{0.24\linewidth}
        \centering
        \includegraphics[width=0.9\linewidth]{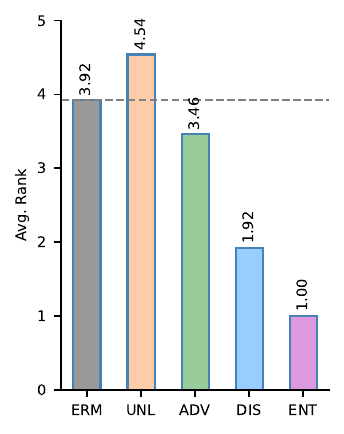}
        \caption{Overall PPV Rank ($\downarrow$)}
        \label{fig:avg-ppv-rank}
    \end{subfigure}
    \begin{subfigure}[b]{0.24\linewidth}
        \centering
        \includegraphics[width=0.9\linewidth]{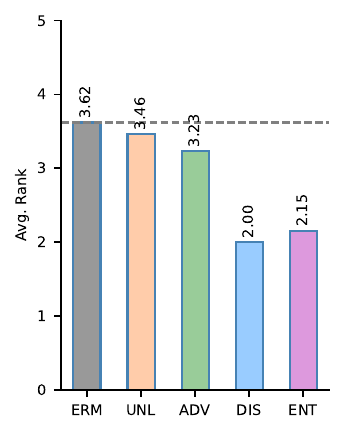}
        \caption{PPV Ratio Rank ($\downarrow$)}
        \label{fig:avg-ppv-ratio-rank}
    \end{subfigure}
    \caption{Average overall precision and group-wise precision ratio (ratio of the best- to worst-performing subgroup) scores and ranks for each method, aggregated over all experiments.}
    \label{fig:ppv-average}
\end{figure*}

\begin{figure*}[ht]
    \centering
    \begin{subfigure}[b]{0.24\linewidth}
        \centering
        \includegraphics[width=0.9\linewidth]{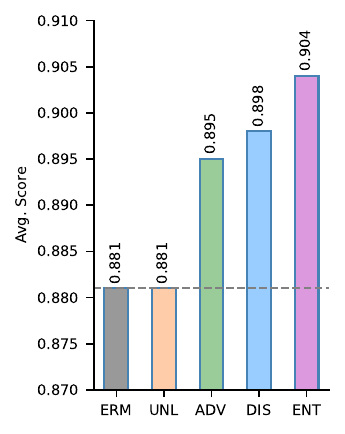}
        \caption{Overall TPR Score ($\uparrow$)}
        \label{fig:avg-tpr}
    \end{subfigure}
    \begin{subfigure}[b]{0.24\linewidth}
        \centering
        \includegraphics[width=0.9\linewidth]{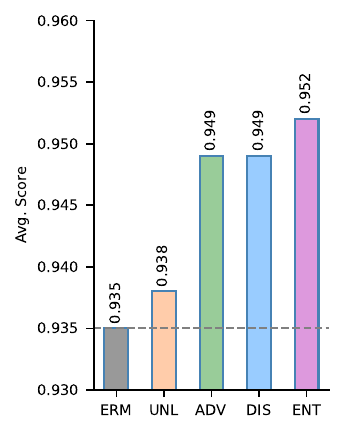}
        \caption{TPR Ratio Score ($\uparrow$)}
        \label{fig:avg-tpr-ratio}
    \end{subfigure}
    \begin{subfigure}[b]{0.24\linewidth}
        \centering
        \includegraphics[width=0.9\linewidth]{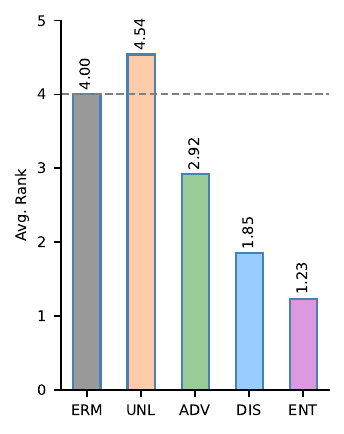}
        \caption{Overall TPR Rank ($\downarrow$)}
        \label{fig:avg-tpr-rank}
    \end{subfigure}
    \begin{subfigure}[b]{0.24\linewidth}
        \centering
        \includegraphics[width=0.9\linewidth]{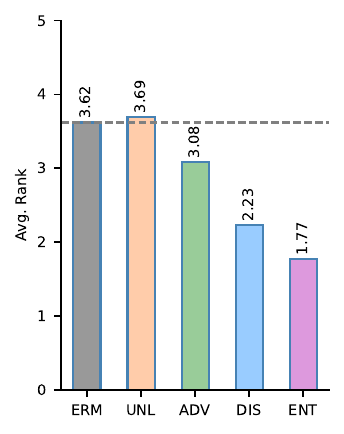}
        \caption{TPR Ratio Rank ($\downarrow$)}
        \label{fig:avg-tpr-ratio-rank}
    \end{subfigure}
    \caption{Average overall recall and group-wise recall ratio (ratio of the best- to worst-performing subgroup) scores and ranks for each method, aggregated over all experiments.}
    \label{fig:tpr-average}
\end{figure*}

\begin{figure*}[ht]
    \centering
    \begin{subfigure}[b]{0.24\linewidth}
        \centering
        \includegraphics[width=0.9\linewidth]{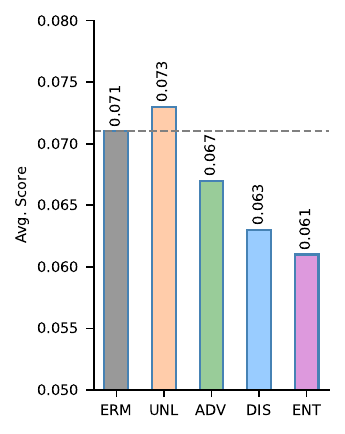}
        \caption{Overall FPR Score ($\uparrow$)}
        \label{fig:avg-fpr}
    \end{subfigure}
    \begin{subfigure}[b]{0.24\linewidth}
        \centering
        \includegraphics[width=0.9\linewidth]{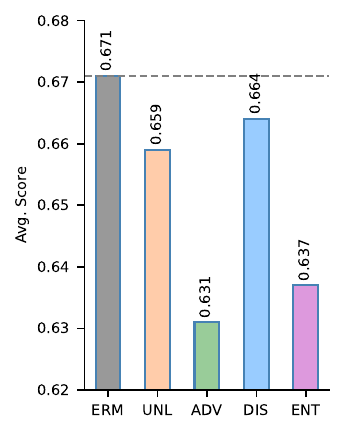}
        \caption{FPR Ratio Score ($\uparrow$)}
        \label{fig:avg-fpr-ratio}
    \end{subfigure}
    \begin{subfigure}[b]{0.24\linewidth}
        \centering
        \includegraphics[width=0.9\linewidth]{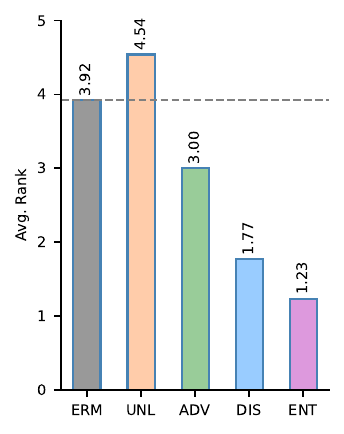}
        \caption{Overall FPR Rank ($\downarrow$)}
        \label{fig:avg-fpr-rank}
    \end{subfigure}
    \begin{subfigure}[b]{0.24\linewidth}
        \centering
        \includegraphics[width=0.9\linewidth]{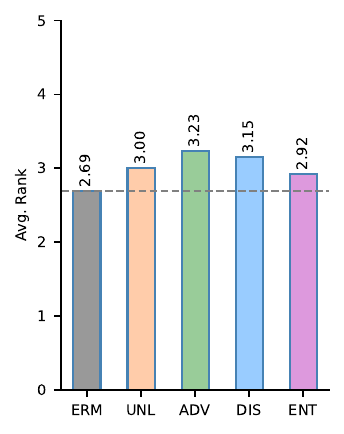}
        \caption{FPR Ratio Rank ($\downarrow$)}
        \label{fig:avg-fpr-ratio-rank}
    \end{subfigure}
    \caption{Average overall FPR and group-wise FPR ratio (ratio of the best- to worst-performing subgroup) scores and ranks for each method, aggregated over all experiments.}
    \label{fig:fpr-average}
\end{figure*}

Table~\ref{table:dataset} summarizes the downstream and sensitive attributes of the datasets used in our experiments. Tables~\ref{tab:performance-utk-fair-water},~\ref{tab:performance-celeba-1},~and~\ref{tab:performance-celeba-2} report overall utility and subgroup-wise ratio/difference for accuracy, precision, recall, and FPR. Figures~\ref{fig:acc-average},~\ref{fig:ppv-average},~\ref{fig:tpr-average},~and~\ref{fig:fpr-average} present method-wise accuracy, precision, recall, and FPR scores and their rankings averaged across datasets (also see Table~\ref{tab:average-table}). Tables~\ref{tab:hyper-auc-performance-utk-age}~and~\ref{tab:hyper-auc-performance-utk-gender} report sensitivity analysis of method-specific hyperparameters.

\begin{table*}[ht]
\centering
\resizebox{0.99\linewidth}{!}{
\begin{tabular}{lll}
\toprule
\textbf{Dataset} & \textbf{Downstream Task Attribute(s)} & \textbf{Sensitive Attribute(s)} \\
\midrule
CelebA & Bald; Black hair; Eyeglasses; Smiling; & Gender (Male, Female) \\
 & Wearing hat; Young; Attractive; Blond hair & \\
\midrule
UTKFace & Age (5 bins) & Race (White, Black, Asian, Indian, and Others) \\
 & Gender (Male, Female) & \\
\midrule 
FairFace & Age (9 bins) & Race (White, Southeast Asian, Middle Eastern, Black, Indian, Latino Hispanic, East Asian) \\
 & Gender (Male, Female) & \\
\midrule
WaterBird & Bird type (Water bird, Land bird) & Background (Water, Land) \\
\bottomrule
\end{tabular}
}
\caption{Summary of downstream labels and sensitive attributes for each dataset, covering both binary and multi-valued cases.}
\label{table:dataset}
\end{table*}

\begin{table*}[ht]
\centering
\resizebox{0.7\linewidth}{!}{
\begin{tabular}{llccccc}
\toprule
\textbf{Metric} & \textbf{Method} & \multicolumn{2}{c}{\textbf{UTKFace}} & \multicolumn{2}{c}{\textbf{FairFace}} & \textbf{WaterBird} \\
\cmidrule(lr){3-4}
\cmidrule(lr){5-6}
\cmidrule(lr){7-7}
& & Age $\times$ Race & Gender $\times$ Race & Age $\times$ Race & Gender $\times$ Race & Type $\times$ Background \\
\midrule
\multirow{5}{*}{\texttt{ACC} ($\uparrow$)}
& \textsc{Erm} & \cellcolor{gray!20}0.757 ± 0.004 & \cellcolor{gray!20}0.949 ± 0.001 & \cellcolor{gray!20}0.532 ± 0.002 & \cellcolor{gray!20}0.934 ± 0.001 & \cellcolor{gray!20}0.953 ± 0.004 \\
& \textsc{Unl} & \textcolor{biasred}{0.750 ± 0.004} & 0.946 ± 0.002 & 0.529 ± 0.004 & \textcolor{biasred}{0.928 ± 0.001} & 0.957 ± 0.001 \\
& \textsc{Adv} & \textcolor{biasgreen}{0.765 ± 0.002} & 0.946 ± 0.003 & 0.533 ± 0.007 & 0.938 ± 0.001 & \textcolor{biasred}{0.935 ± 0.012} \\
& \textsc{Dis} & \textcolor{biasgreen}{0.767 ± 0.002} & 0.954 ± 0.001 & \textcolor{biasgreen}{0.546 ± 0.004} & \textcolor{biasgreen}{0.942 ± 0.001} & \textcolor{biasgreen}{0.961 ± 0.004} \\
& \textsc{Ent} & \textcolor{biasgreen}{\textbf{0.778 ± 0.006}} & \textcolor{biasgreen}{\textbf{0.956 ± 0.001}} & \textcolor{biasgreen}{\textbf{0.553 ± 0.005}} & \textcolor{biasgreen}{\textbf{0.943 ± 0.000}} & \textcolor{biasgreen}{\textbf{0.971 ± 0.002}} \\
\midrule
\multirow{5}{*}{$\Delta$\texttt{ACC} ($\downarrow$)}
& \textsc{Erm} & \cellcolor{gray!20}0.097 ± 0.006 & \cellcolor{gray!20}0.063 ± 0.002 & \cellcolor{gray!20}0.055 ± 0.001 & \cellcolor{gray!20}0.075 ± 0.005 & \cellcolor{gray!20}0.036 ± 0.003 \\
& \textsc{Unl} & \textcolor{biasred}{0.108 ± 0.005} & \textcolor{biasgreen}{\textbf{0.046 ± 0.008}} & \textcolor{biasred}{0.071 ± 0.012} & 0.077 ± 0.006 & 0.035 ± 0.004 \\
& \textsc{Adv} & \textcolor{biasgreen}{\textbf{0.072 ± 0.014}} & 0.060 ± 0.004 & \textcolor{biasred}{0.068 ± 0.006} & 0.073 ± 0.005 & \textcolor{biasred}{0.044 ± 0.026} \\
& \textsc{Dis} & \textcolor{biasgreen}{0.073 ± 0.007} & 0.061 ± 0.002 & \textbf{0.053 ± 0.004} & \textcolor{biasgreen}{0.066 ± 0.004} & \textcolor{biasgreen}{0.027 ± 0.003} \\
& \textsc{Ent} & \textcolor{biasgreen}{0.086 ± 0.012} & \textcolor{biasgreen}{0.056 ± 0.004} & 0.054 ± 0.004 & \textcolor{biasgreen}{\textbf{0.065 ± 0.004}} & \textcolor{biasgreen}{\textbf{0.009 ± 0.005}} \\
\midrule
\multirow{5}{*}{\texttt{ACC} ratio ($\uparrow$)}
& \textsc{Erm} & \cellcolor{gray!20}0.883 ± 0.007 & \cellcolor{gray!20}0.934 ± 0.002 & \cellcolor{gray!20}0.903 ± 0.002 & \cellcolor{gray!20}0.922 ± 0.005 & \cellcolor{gray!20}0.963 ± 0.003 \\
& \textsc{Unl} & \textcolor{biasred}{0.868 ± 0.007} & \textcolor{biasgreen}{\textbf{0.953 ± 0.008}} & \textcolor{biasred}{0.873 ± 0.020} & 0.920 ± 0.006 & 0.964 ± 0.004 \\
& \textsc{Adv} & \textcolor{biasgreen}{0.911 ± 0.017} & 0.937 ± 0.004 & \textcolor{biasred}{0.881 ± 0.011} & 0.924 ± 0.005 & \textcolor{biasred}{0.954 ± 0.027} \\
& \textsc{Dis} & \textcolor{biasgreen}{\textbf{0.911 ± 0.008}} & 0.937 ± 0.002 & \textbf{0.907 ± 0.007} & \textcolor{biasgreen}{0.932 ± 0.004} & \textcolor{biasgreen}{0.972 ± 0.003} \\
& \textsc{Ent} & \textcolor{biasgreen}{0.897 ± 0.013} & \textcolor{biasgreen}{0.942 ± 0.004} & \textbf{0.907 ± 0.007} & \textcolor{biasgreen}{\textbf{0.933 ± 0.004}} & \textcolor{biasgreen}{\textbf{0.990 ± 0.005}} \\
\midrule
\midrule
\multirow{5}{*}{\texttt{PPV} ($\uparrow$)}
& \textsc{Erm} & \cellcolor{gray!20}0.625 ± 0.025 & \cellcolor{gray!20}0.949 ± 0.001 & \cellcolor{gray!20}0.502 ± 0.004 & \cellcolor{gray!20}0.933 ± 0.001 & \cellcolor{gray!20}0.926 ± 0.006 \\
& \textsc{Unl} & \textcolor{biasgreen}{0.642 ± 0.025} & 0.946 ± 0.002 & 0.499 ± 0.005 & 0.928 ± 0.001 & \textcolor{biasgreen}{0.934 ± 0.002} \\
& \textsc{Adv} & \textcolor{biasgreen}{0.687 ± 0.019} & 0.946 ± 0.003 & \textcolor{biasgreen}{0.513 ± 0.006} & 0.938 ± 0.001 & \textcolor{biasred}{0.896 ± 0.015} \\
& \textsc{Dis} & \textcolor{biasgreen}{0.715 ± 0.009} & 0.954 ± 0.001 & \textcolor{biasgreen}{0.525 ± 0.005} & \textcolor{biasgreen}{0.942 ± 0.001} & \textcolor{biasgreen}{0.938 ± 0.006} \\
& \textsc{Ent} & \textcolor{biasgreen}{\textbf{0.751 ± 0.023}} & \textcolor{biasgreen}{\textbf{0.956 ± 0.001}} & \textcolor{biasgreen}{\textbf{0.547 ± 0.004}} & \textcolor{biasgreen}{\textbf{0.943 ± 0.000}} & \textcolor{biasgreen}{\textbf{0.952 ± 0.003}} \\
\midrule
\multirow{5}{*}{$\Delta$\texttt{PPV} ($\downarrow$)}
& \textsc{Erm} & \cellcolor{gray!20}0.212 ± 0.061 & \cellcolor{gray!20}0.062 ± 0.002 & \cellcolor{gray!20}0.099 ± 0.007 & \cellcolor{gray!20}0.065 ± 0.006 & \cellcolor{gray!20}0.018 ± 0.008 \\
& \textsc{Unl} & \textcolor{biasred}{0.236 ± 0.039} & \textcolor{biasgreen}{\textbf{0.044 ± 0.008}} & 0.095 ± 0.004 & 0.069 ± 0.006 & \textcolor{biasgreen}{\textbf{0.011 ± 0.004}} \\
& \textsc{Adv} & \textcolor{biasgreen}{\textbf{0.147 ± 0.067}} & 0.059 ± 0.005 & 0.102 ± 0.012 & 0.064 ± 0.006 & \textcolor{biasred}{0.041 ± 0.020} \\
& \textsc{Dis} & \textcolor{biasgreen}{0.169 ± 0.051} & 0.060 ± 0.002 & \textcolor{biasgreen}{0.091 ± 0.013} & \textcolor{biasgreen}{\textbf{0.057 ± 0.005}} & 0.020 ± 0.006 \\
& \textsc{Ent} & \textcolor{biasgreen}{0.177 ± 0.055} & \textcolor{biasgreen}{0.054 ± 0.003} & \textcolor{biasgreen}{\textbf{0.089 ± 0.008}} & \textcolor{biasgreen}{\textbf{0.057 ± 0.005}} & \textcolor{biasred}{0.027 ± 0.013} \\
\midrule
\multirow{5}{*}{\texttt{PPV} ratio ($\uparrow$)}
& \textsc{Erm} & \cellcolor{gray!20}0.728 ± 0.071 & \cellcolor{gray!20}0.935 ± 0.002 & \cellcolor{gray!20}0.821 ± 0.010 & \cellcolor{gray!20}0.932 ± 0.006 & \cellcolor{gray!20}0.981 ± 0.009 \\
& \textsc{Unl} & \textcolor{biasred}{0.683 ± 0.037} & \textcolor{biasgreen}{\textbf{0.954 ± 0.008}} & 0.826 ± 0.006 & 0.927 ± 0.007 & \textcolor{biasgreen}{\textbf{0.988 ± 0.005}} \\
& \textsc{Adv} & \textcolor{biasgreen}{\textbf{0.810 ± 0.078}} & 0.938 ± 0.005 & 0.822 ± 0.019 & 0.933 ± 0.006 & \textcolor{biasred}{0.955 ± 0.020} \\
& \textsc{Dis} & \textcolor{biasgreen}{0.786 ± 0.064} & 0.938 ± 0.002 & \textcolor{biasgreen}{0.843 ± 0.019} & \textcolor{biasgreen}{\textbf{0.941 ± 0.005}} & 0.979 ± 0.006 \\
& \textsc{Ent} & \textcolor{biasgreen}{0.779 ± 0.064} & \textcolor{biasgreen}{0.944 ± 0.003} & \textcolor{biasgreen}{\textbf{0.850 ± 0.012}} & \textcolor{biasgreen}{\textbf{0.941 ± 0.005}} & \textcolor{biasred}{0.972 ± 0.013} \\
\midrule
\midrule
\multirow{5}{*}{\texttt{TPR} ($\uparrow$)}
& \textsc{Erm} & \cellcolor{gray!20}0.662 ± 0.025 & \cellcolor{gray!20}0.949 ± 0.001 & \cellcolor{gray!20}0.582 ± 0.001 & \cellcolor{gray!20}0.934 ± 0.001 & \cellcolor{gray!20}0.944 ± 0.002 \\
& \textsc{Unl} & \textcolor{biasgreen}{0.681 ± 0.020} & 0.947 ± 0.002 & 0.581 ± 0.002 & 0.929 ± 0.001 & 0.945 ± 0.002 \\
& \textsc{Adv} & \textcolor{biasgreen}{0.768 ± 0.006} & 0.946 ± 0.003 & \textcolor{biasgreen}{0.606 ± 0.002} & 0.938 ± 0.001 & \textcolor{biasred}{0.935 ± 0.010} \\
& \textsc{Dis} & \textcolor{biasgreen}{0.755 ± 0.011} & 0.954 ± 0.001 & \textcolor{biasgreen}{0.609 ± 0.004} & \textcolor{biasgreen}{0.942 ± 0.001} & \textcolor{biasgreen}{0.956 ± 0.001} \\
& \textsc{Ent} & \textcolor{biasgreen}{\textbf{0.789 ± 0.003}} & \textcolor{biasgreen}{\textbf{0.956 ± 0.001}} & \textcolor{biasgreen}{\textbf{0.627 ± 0.003}} & \textcolor{biasgreen}{\textbf{0.943 ± 0.000}} & \textcolor{biasgreen}{\textbf{0.969 ± 0.002}} \\
\midrule
\multirow{5}{*}{$\Delta$\texttt{TPR} ($\downarrow$)}
& \textsc{Erm} & \cellcolor{gray!20}0.270 ± 0.053 & \cellcolor{gray!20}0.064 ± 0.003 & \cellcolor{gray!20}0.083 ± 0.014 & \cellcolor{gray!20}0.073 ± 0.005 & \cellcolor{gray!20}\textbf{0.007 ± 0.002} \\
& \textsc{Unl} & \textcolor{biasgreen}{0.215 ± 0.018} & \textcolor{biasgreen}{\textbf{0.047 ± 0.007}} & 0.079 ± 0.012 & 0.073 ± 0.006 & \textcolor{biasred}{0.013 ± 0.000} \\
& \textsc{Adv} & \textcolor{biasgreen}{\textbf{0.161 ± 0.024}} & 0.062 ± 0.004 & \textcolor{biasgreen}{0.066 ± 0.002} & 0.069 ± 0.005 & \textcolor{biasred}{0.021 ± 0.006} \\
& \textsc{Dis} & \textcolor{biasgreen}{0.205 ± 0.009} & 0.062 ± 0.002 & \textcolor{biasgreen}{\textbf{0.064 ± 0.015}} & \textcolor{biasgreen}{\textbf{0.063 ± 0.004}} & 0.008 ± 0.007 \\
& \textsc{Ent} & \textcolor{biasgreen}{0.168 ± 0.018} & \textcolor{biasgreen}{0.058 ± 0.005} & \textcolor{biasgreen}{0.076 ± 0.009} & \textcolor{biasgreen}{\textbf{0.063 ± 0.004}} & 0.012 ± 0.004 \\
\midrule
\multirow{5}{*}{\texttt{TPR} ratio ($\uparrow$)}
& \textsc{Erm} & \cellcolor{gray!20}0.682 ± 0.050 & \cellcolor{gray!20}0.933 ± 0.003 & \cellcolor{gray!20}0.865 ± 0.021 & \cellcolor{gray!20}0.924 ± 0.006 & \cellcolor{gray!20}\textbf{0.993 ± 0.003} \\
& \textsc{Unl} & \textcolor{biasgreen}{0.728 ± 0.032} & \textcolor{biasgreen}{\textbf{0.951 ± 0.007}} & \textcolor{biasgreen}{0.871 ± 0.018} & 0.923 ± 0.007 & \textcolor{biasred}{0.986 ± 0.000} \\
& \textsc{Adv} & \textcolor{biasgreen}{\textbf{0.809 ± 0.027}} & 0.936 ± 0.004 & \textcolor{biasgreen}{0.896 ± 0.003} & 0.928 ± 0.005 & \textcolor{biasred}{0.977 ± 0.006} \\
& \textsc{Dis} & \textcolor{biasgreen}{0.761 ± 0.013} & 0.936 ± 0.002 & \textcolor{biasgreen}{\textbf{0.901 ± 0.023}} & \textcolor{biasgreen}{\textbf{0.935 ± 0.004}} & 0.991 ± 0.007 \\
& \textsc{Ent} & \textcolor{biasgreen}{0.808 ± 0.020} & \textcolor{biasgreen}{0.940 ± 0.005} & \textcolor{biasgreen}{0.885 ± 0.013} & \textcolor{biasgreen}{\textbf{0.935 ± 0.004}} & \textcolor{biasred}{0.987 ± 0.004} \\
\midrule
\midrule
\multirow{5}{*}{\texttt{FPR} ($\downarrow$)}
& \textsc{Erm} & \cellcolor{gray!20}0.073 ± 0.001 & \cellcolor{gray!20}0.051 ± 0.001 & \cellcolor{gray!20}0.063 ± 0.000 & \cellcolor{gray!20}0.066 ± 0.001 & \cellcolor{gray!20}0.056 ± 0.002 \\
& \textsc{Unl} & 0.076 ± 0.001 & 0.053 ± 0.002 & 0.063 ± 0.001 & 0.071 ± 0.001 & 0.055 ± 0.002 \\
& \textsc{Adv} & 0.072 ± 0.000 & 0.054 ± 0.003 & 0.062 ± 0.001 & 0.062 ± 0.001 & \textcolor{biasred}{0.065 ± 0.010} \\
& \textsc{Dis} & 0.071 ± 0.001 & 0.046 ± 0.001 & 0.061 ± 0.000 & \textcolor{biasgreen}{0.058 ± 0.001} & \textcolor{biasgreen}{0.044 ± 0.001} \\
& \textsc{Ent} & \textcolor{biasgreen}{\textbf{0.067 ± 0.002}} & \textcolor{biasgreen}{\textbf{0.044 ± 0.001}} & \textbf{0.060 ± 0.000} & \textcolor{biasgreen}{\textbf{0.057 ± 0.000}} & \textcolor{biasgreen}{\textbf{0.031 ± 0.002}} \\
\midrule
\multirow{5}{*}{$\Delta$\texttt{FPR} ($\downarrow$)}
& \textsc{Erm} & \cellcolor{gray!20}0.042 ± 0.010 & \cellcolor{gray!20}0.064 ± 0.003 & \cellcolor{gray!20}0.006 ± 0.000 & \cellcolor{gray!20}0.073 ± 0.005 & \cellcolor{gray!20}\textbf{0.007 ± 0.002} \\
& \textsc{Unl} &  \textcolor{biasred}{0.053 ± 0.005} & \textcolor{biasgreen}{\textbf{0.047 ± 0.007}} & 0.008 ± 0.002 & 0.073 ± 0.006 & \textcolor{biasred}{0.013 ± 0.000} \\
& \textsc{Adv} & 0.044 ± 0.010 & 0.062 ± 0.004 & 0.007 ± 0.001 & 0.069 ± 0.005 & \textcolor{biasred}{0.021 ± 0.006} \\
& \textsc{Dis} & 0.043 ± 0.011 & 0.062 ± 0.002 & 0.006 ± 0.000 & \textcolor{biasgreen}{\textbf{0.063 ± 0.004}} & 0.008 ± 0.007 \\
& \textsc{Ent} & \textbf{0.041 ± 0.007} & \textcolor{biasgreen}{0.058 ± 0.005} & \textbf{0.005 ± 0.001} & \textcolor{biasgreen}{\textbf{0.063 ± 0.004}} & 0.012 ± 0.004 \\
\midrule
\multirow{5}{*}{\texttt{FPR} ratio ($\uparrow$)}
& \textsc{Erm} & \cellcolor{gray!20}\textbf{0.631 ± 0.055} & \cellcolor{gray!20}0.364 ± 0.019 & \cellcolor{gray!20}0.909 ± 0.006 & \cellcolor{gray!20}0.343 ± 0.024 & \cellcolor{gray!20}\textbf{0.907 ± 0.032} \\
& \textsc{Unl} & \textcolor{biasred}{0.560 ± 0.030} & \textcolor{biasgreen}{\textbf{0.461 ± 0.039}} & \textcolor{biasred}{0.885 ± 0.023} & \textcolor{biasgreen}{\textbf{0.393 ± 0.034}} & \textcolor{biasred}{0.810 ± 0.005} \\
& \textsc{Adv} & \textcolor{biasred}{0.606 ± 0.054} & \textcolor{biasgreen}{0.399 ± 0.009} & \textcolor{biasred}{0.889 ± 0.009} & \textcolor{biasgreen}{0.360 ± 0.025} & \textcolor{biasred}{0.772 ± 0.061} \\
& \textsc{Dis} & \textcolor{biasred}{0.611 ± 0.059} & \textcolor{biasred}{0.340 ± 0.015} & \textcolor{biasred}{0.901 ± 0.001} & \textcolor{biasgreen}{0.367 ± 0.024} & \textcolor{biasred}{0.876 ± 0.094} \\
& \textsc{Ent} & \textcolor{biasred}{0.599 ± 0.046} & \textcolor{biasred}{0.337 ± 0.036} & \textcolor{biasgreen}{\textbf{0.915 ± 0.009}} & 0.344 ± 0.023 & \textcolor{biasred}{0.731 ± 0.059} \\
\bottomrule
\end{tabular}
}
\caption{Performance evaluation of fine-tuning strategies on UTKFace, FairFace, and WaterBird datasets, averaged over three random seeds. The best values are shown in \textbf{bold}. The scores that improve upon \textsc{Erm} by more than 0.005 are highlighted in \textcolor{biasgreen}{green}, while degradations greater than 0.005 are shown in \textcolor{biasred}{red}. Changes within ±0.005 of \textsc{Erm} score are considered insignificant.}
\label{tab:performance-utk-fair-water}
\end{table*}

\begin{table*}[ht]
\centering
\resizebox{0.68\linewidth}{!}{
\begin{tabular}{llcccc}
\toprule
\textbf{Metric} & \textbf{Method} & \multicolumn{4}{c}{\textbf{CelebA}} \\
\cmidrule(lr){3-6}
& & Bald $\times$ Gender & Black hair $\times$ Gender & Eyeglasses $\times$ Gender & Smiling $\times$ Gender \\
\midrule
\multirow{5}{*}{\texttt{ACC} ($\uparrow$)}
& \textsc{Erm} & \cellcolor{gray!20} 0.986 ± 0.000 & \cellcolor{gray!20} 0.891 ± 0.000 & \cellcolor{gray!20} 0.995 ± 0.000 & \cellcolor{gray!20} 0.928 ± 0.000 \\
& \textsc{Unl} & 0.984 ± 0.000 & \textcolor{biasred}{0.885 ± 0.001} & 0.994 ± 0.000 & 0.925 ± 0.002 \\
& \textsc{Adv} & 0.983 ± 0.002 & 0.892 ± 0.005 & 0.996 ± 0.000 & 0.930 ± 0.001 \\
& \textsc{Dis} & 0.990 ± 0.000 & 0.894 ± 0.002 & \textbf{0.997 ± 0.000} & \textbf{0.933 ± 0.001} \\
& \textsc{Ent} & \textbf{0.991 ± 0.000} & \textbf{0.896 ± 0.004} & \textbf{0.997 ± 0.000} & 0.933 ± 0.002 \\
\midrule
\multirow{5}{*}{$\Delta$\texttt{ACC} ($\downarrow$)}
& \textsc{Erm} & \cellcolor{gray!20} 0.034 ± 0.001 & \cellcolor{gray!20} \textbf{0.039 ± 0.001} & \cellcolor{gray!20} 0.007 ± 0.001 & \cellcolor{gray!20} 0.013 ± 0.000 \\
& \textsc{Unl} & 0.038 ± 0.000 & 0.042 ± 0.001 & 0.008 ± 0.000 & 0.015 ± 0.001 \\
& \textsc{Adv} & 0.039 ± 0.004 & 0.039 ± 0.003 & 0.007 ± 0.000 & \textbf{0.012 ± 0.001} \\
& \textsc{Dis} & \textcolor{biasgreen}{0.025 ± 0.001} & 0.039 ± 0.002 & 0.005 ± 0.000 & 0.013 ± 0.001 \\
& \textsc{Ent} & \textcolor{biasgreen}{\textbf{0.021 ± 0.001}} & 0.039 ± 0.003 & \textbf{0.004 ± 0.000} & \textbf{0.012 ± 0.001} \\
\midrule
\multirow{5}{*}{\texttt{ACC} ratio ($\uparrow$)}
& \textsc{Erm} & \cellcolor{gray!20} 0.966 ± 0.001 & \cellcolor{gray!20} \textbf{0.957 ± 0.001} & \cellcolor{gray!20} 0.993 ± 0.001 & \cellcolor{gray!20} 0.986 ± 0.000 \\
& \textsc{Unl} & 0.962 ± 0.000 & 0.954 ± 0.001 & 0.992 ± 0.000 & 0.984 ± 0.001 \\
& \textsc{Adv} & 0.961 ± 0.004 & 0.957 ± 0.004 & 0.993 ± 0.000 & \textbf{0.987 ± 0.001} \\
& \textsc{Dis} & \textcolor{biasgreen}{0.975 ± 0.001} & 0.957 ± 0.002 & 0.995 ± 0.000 & 0.986 ± 0.001 \\
& \textsc{Ent} & \textcolor{biasgreen}{\textbf{0.979 ± 0.001}} & 0.957 ± 0.003 & \textbf{0.996 ± 0.000} & \textbf{0.987 ± 0.001} \\
\midrule
\midrule
\multirow{5}{*}{\texttt{PPV} ($\uparrow$)}
& \textsc{Erm} & \cellcolor{gray!20} 0.808 ± 0.004 & \cellcolor{gray!20} 0.841 ± 0.001 & \cellcolor{gray!20} 0.972 ± 0.002 & \cellcolor{gray!20} 0.928 ± 0.000 \\
& \textsc{Unl} & \textcolor{biasred}{0.793 ± 0.001} & \textcolor{biasred}{0.834 ± 0.001} & 0.968 ± 0.001 & 0.926 ± 0.002 \\
& \textsc{Adv} & \textcolor{biasred}{0.788 ± 0.014} & 0.842 ± 0.006 & 0.971 ± 0.000 & 0.930 ± 0.001 \\
& \textsc{Dis} & \textcolor{biasgreen}{0.844 ± 0.004} & 0.844 ± 0.002 & \textcolor{biasgreen}{0.979 ± 0.002} & \textbf{0.933 ± 0.001} \\
& \textsc{Ent} & \textcolor{biasgreen}{\textbf{0.860 ± 0.005}} & \textcolor{biasgreen}{\textbf{0.847 ± 0.004}} & \textcolor{biasgreen}{\textbf{0.985 ± 0.000}} & 0.933 ± 0.002 \\
\midrule
\multirow{5}{*}{$\Delta$\texttt{PPV} ($\downarrow$)}
& \textsc{Erm} & \cellcolor{gray!20} 0.102 ± 0.021 & \cellcolor{gray!20} 0.002 ± 0.000 & \cellcolor{gray!20} 0.004 ± 0.001 & \cellcolor{gray!20} 0.015 ± 0.000 \\
& \textsc{Unl} & 0.101 ± 0.059 & \textbf{0.001 ± 0.000} & 0.005 ± 0.003 & 0.016 ± 0.001 \\
& \textsc{Adv} & \textcolor{biasred}{0.129 ± 0.036} & 0.003 ± 0.001 & \textbf{0.001 ± 0.001} & \textbf{0.013 ± 0.002} \\
& \textsc{Dis} & \textcolor{biasgreen}{\textbf{0.027 ± 0.006}} & 0.004 ± 0.002 & \textbf{0.001 ± 0.001} & 0.014 ± 0.001 \\
& \textsc{Ent} & \textcolor{biasgreen}{0.041 ± 0.030} & 0.007 ± 0.001 & 0.002 ± 0.001 & 0.015 ± 0.001 \\
\midrule
\multirow{5}{*}{\texttt{PPV} ratio ($\uparrow$)}
& \textsc{Erm} & \cellcolor{gray!20} 0.874 ± 0.026 & \cellcolor{gray!20} 0.997 ± 0.001 & \cellcolor{gray!20} 0.996 ± 0.001 & \cellcolor{gray!20} 0.984 ± 0.000 \\
& \textsc{Unl} & 0.872 ± 0.075 & \textbf{0.999 ± 0.000} & 0.994 ± 0.003 & 0.982 ± 0.001 \\
& \textsc{Adv} & \textcolor{biasred}{0.836 ± 0.047} & 0.996 ± 0.002 & \textbf{0.999 ± 0.001} & \textbf{0.986 ± 0.002} \\
& \textsc{Dis} & \textcolor{biasgreen}{\textbf{0.968 ± 0.008}} & 0.996 ± 0.002 & \textbf{0.999 ± 0.001} & 0.985 ± 0.001 \\
& \textsc{Ent} & \textcolor{biasgreen}{0.952 ± 0.036} & \textcolor{biasred}{0.991 ± 0.001} & 0.998 ± 0.001 & 0.984 ± 0.001 \\
\midrule
\midrule
\multirow{5}{*}{\texttt{TPR} ($\uparrow$)}
& \textsc{Erm} & \cellcolor{gray!20} 0.957 ± 0.003 & \cellcolor{gray!20} 0.901 ± 0.001 & \cellcolor{gray!20} 0.988 ± 0.001 & \cellcolor{gray!20} 0.927 ± 0.001 \\
& \textsc{Unl} & \textcolor{biasred}{0.949 ± 0.006} & 0.898 ± 0.001 & 0.988 ± 0.001 & 0.925 ± 0.002 \\
& \textsc{Adv} & \textcolor{biasgreen}{\textbf{0.985 ± 0.001}} & 0.904 ± 0.001 & \textcolor{biasgreen}{0.994 ± 0.000} & 0.929 ± 0.001 \\
& \textsc{Dis} & \textcolor{biasgreen}{0.983 ± 0.001} & \textcolor{biasgreen}{\textbf{0.907 ± 0.001}} & \textcolor{biasgreen}{\textbf{0.995 ± 0.000}} & \textcolor{biasgreen}{\textbf{0.933 ± 0.001}} \\
& \textsc{Ent} & \textcolor{biasgreen}{0.980 ± 0.002} & \textcolor{biasgreen}{\textbf{0.907 ± 0.001}} & \textcolor{biasgreen}{0.994 ± 0.000} & \textcolor{biasgreen}{0.933 ± 0.002} \\
\midrule
\multirow{5}{*}{$\Delta$\texttt{TPR} ($\downarrow$)}
& \textsc{Erm} & \cellcolor{gray!20} 0.076 ± 0.019 & \cellcolor{gray!20} 0.029 ± 0.002 & \cellcolor{gray!20} 0.005 ± 0.000 & \cellcolor{gray!20} 0.018 ± 0.001 \\
& \textsc{Unl} & \textcolor{biasred}{0.097 ± 0.041} & 0.032 ± 0.002 & 0.005 ± 0.002 & 0.021 ± 0.001 \\
& \textsc{Adv} & \textcolor{biasgreen}{0.053 ± 0.025} & 0.029 ± 0.005 & 0.006 ± 0.002 & 0.018 ± 0.003 \\
& \textsc{Dis} & \textcolor{biasgreen}{\textbf{0.050 ± 0.025}} & 0.029 ± 0.001 & \textbf{0.001 ± 0.000} & 0.020 ± 0.001 \\
& \textsc{Ent} & \textcolor{biasgreen}{0.051 ± 0.025} & \textbf{0.028 ± 0.000} & 0.004 ± 0.001 & \textbf{0.016 ± 0.002} \\
\midrule
\multirow{5}{*}{\texttt{TPR} ratio ($\uparrow$)}
& \textsc{Erm} & \cellcolor{gray!20} 0.920 ± 0.021 & \cellcolor{gray!20} 0.968 ± 0.002 & \cellcolor{gray!20} 0.995 ± 0.000 & \cellcolor{gray!20} 0.980 ± 0.002 \\
& \textsc{Unl} & \textcolor{biasred}{0.898 ± 0.046} & 0.965 ± 0.002 & 0.995 ± 0.002 & 0.978 ± 0.001 \\
& \textsc{Adv} & \textcolor{biasgreen}{0.946 ± 0.025} & 0.968 ± 0.006 & 0.994 ± 0.002 & 0.980 ± 0.004 \\
& \textsc{Dis} & \textcolor{biasgreen}{\textbf{0.949 ± 0.026}} & 0.969 ± 0.001 & \textbf{0.999 ± 0.000} & 0.979 ± 0.001 \\
& \textsc{Ent} & \textcolor{biasgreen}{0.948 ± 0.026} & \textbf{0.970 ± 0.000} & 0.996 ± 0.001 & \textbf{0.983 ± 0.002} \\
\midrule
\midrule
\multirow{5}{*}{\texttt{FPR} ($\downarrow$)}
& \textsc{Erm} & \cellcolor{gray!20} 0.043 ± 0.003 & \cellcolor{gray!20} 0.099 ± 0.001 & \cellcolor{gray!20} 0.012 ± 0.001 & \cellcolor{gray!20} 0.073 ± 0.001 \\
& \textsc{Unl} & \textcolor{biasred}{0.051 ± 0.006} & 0.102 ± 0.001 & 0.012 ± 0.001 & 0.075 ± 0.002 \\
& \textsc{Adv} & \textcolor{biasgreen}{\textbf{0.015 ± 0.001}} & 0.096 ± 0.001 & \textcolor{biasgreen}{0.006 ± 0.000} & 0.071 ± 0.001 \\
& \textsc{Dis} & \textcolor{biasgreen}{0.017 ± 0.001} & \textcolor{biasgreen}{\textbf{0.093 ± 0.001 }}& \textcolor{biasgreen}{\textbf{0.005 ± 0.000}} & \textcolor{biasgreen}{\textbf{0.067 ± 0.001}} \\
& \textsc{Ent} & \textcolor{biasgreen}{0.020 ± 0.002} & \textcolor{biasgreen}{\textbf{0.093 ± 0.001}} & \textcolor{biasgreen}{0.006 ± 0.000} & \textcolor{biasgreen}{0.067 ± 0.002} \\
\midrule
\multirow{5}{*}{$\Delta$\texttt{FPR} ($\downarrow$)}
& \textsc{Erm} & \cellcolor{gray!20} 0.076 ± 0.019 & \cellcolor{gray!20} 0.029 ± 0.002 & \cellcolor{gray!20} 0.005 ± 0.000 & \cellcolor{gray!20} 0.018 ± 0.001 \\
& \textsc{Unl} & \textcolor{biasred}{0.097 ± 0.041} & 0.032 ± 0.002 & 0.005 ± 0.002 & 0.021 ± 0.001 \\
& \textsc{Adv} & \textcolor{biasgreen}{0.053 ± 0.025} & 0.029 ± 0.005 & 0.006 ± 0.002 & 0.018 ± 0.003 \\
& \textsc{Dis} & \textcolor{biasgreen}{\textbf{0.050 ± 0.025}} & 0.029 ± 0.001 & \textbf{0.001 ± 0.000} & 0.020 ± 0.001 \\
& \textsc{Ent} & \textcolor{biasgreen}{0.051 ± 0.025} & \textbf{0.028 ± 0.000} & 0.004 ± 0.001 & \textbf{0.016 ± 0.002} \\
\midrule
\multirow{5}{*}{\texttt{FPR} ratio ($\uparrow$)}
& \textsc{Erm} & \cellcolor{gray!20} \textbf{0.258 ± 0.132} & \cellcolor{gray!20} \textbf{0.752 ± 0.017} & \cellcolor{gray!20} 0.645 ± 0.010 & \cellcolor{gray!20} 0.786 ± 0.014 \\
& \textsc{Unl} & \textcolor{biasred}{0.096 ± 0.093} & \textcolor{biasred}{0.740 ± 0.012} & \textcolor{biasgreen}{0.682 ± 0.135} & \textcolor{biasred}{0.766 ± 0.010} \\
& \textsc{Adv} & \textcolor{biasred}{0.067 ± 0.059} & \textcolor{biasred}{0.746 ± 0.041} & \textcolor{biasred}{0.358 ± 0.157} & 0.782 ± 0.032 \\
& \textsc{Dis} & \textcolor{biasred}{0.064 ± 0.063} & \textcolor{biasred}{0.742 ± 0.009} & \textcolor{biasgreen}{\textbf{0.781 ± 0.066}} & \textcolor{biasred}{0.754 ± 0.008} \\
& \textsc{Ent} & \textcolor{biasred}{0.065 ± 0.063} & 0.751 ± 0.005 & \textcolor{biasred}{0.475 ± 0.138} & \textcolor{biasgreen}{\textbf{0.793 ± 0.022}} \\
\bottomrule
\end{tabular}
}
\caption{Performance evaluation of fine-tuning strategies on CelebA dataset, averaged over three random seeds. The best values are shown in \textbf{bold}. The scores that improve upon \textsc{Erm} by more than 0.005 are highlighted in \textcolor{biasgreen}{green}, while degradations greater than 0.005 are shown in \textcolor{biasred}{red}. Changes within ±0.005 of the \textsc{Erm} score are considered insignificant.}
\label{tab:performance-celeba-1}
\end{table*}

\begin{table*}[ht]
\centering
\resizebox{0.7\linewidth}{!}{
\begin{tabular}{llcccc}
\toprule
\textbf{Metric} & \textbf{Method} & \multicolumn{4}{c}{\textbf{CelebA}} \\
\cmidrule(lr){3-6}
& & Wearing hat $\times$ Gender & Young $\times$ Gender & Attractive $\times$ Gender & Blond hair $\times$ Gender \\
\midrule
\multirow{5}{*}{\texttt{ACC} ($\uparrow$)}
& \textsc{Erm} & \cellcolor{gray!20} 0.986 ± 0.000 & \cellcolor{gray!20} 0.868 ± 0.001 & \cellcolor{gray!20} 0.827 ± 0.001 & \cellcolor{gray!20} 0.940 ± 0.001 \\
& \textsc{Unl} & 0.987 ± 0.000 & 0.867 ± 0.001 & 0.826 ± 0.000 & 0.937 ± 0.001 \\
& \textsc{Adv} & 0.988 ± 0.001 & 0.869 ± 0.003 & 0.828 ± 0.001 & 0.941 ± 0.003 \\
& \textsc{Dis} & 0.990 ± 0.000 & \textcolor{biasgreen}{0.876 ± 0.001} & 0.831 ± 0.001 & 0.942 ± 0.002 \\
& \textsc{Ent} & \textbf{0.991 ± 0.001} & \textcolor{biasgreen}{\textbf{0.878 ± 0.005}} & \textbf{0.832 ± 0.001} & \textcolor{biasgreen}{\textbf{0.947 ± 0.002}} \\
\midrule
\multirow{5}{*}{$\Delta$\texttt{ACC} ($\downarrow$)}
& \textsc{Erm} & \cellcolor{gray!20} 0.008 ± 0.001 & \cellcolor{gray!20} 0.079 ± 0.002 & \cellcolor{gray!20} \textbf{0.011 ± 0.003} & \cellcolor{gray!20} 0.057 ± 0.000 \\
& \textsc{Unl} & 0.007 ± 0.001 & 0.080 ± 0.002 & 0.015 ± 0.002 & 0.060 ± 0.002 \\
& \textsc{Adv} & 0.007 ± 0.000 & \textbf{0.075 ± 0.002} & 0.013 ± 0.003 & 0.058 ± 0.003 \\
& \textsc{Dis} & \textbf{0.006 ± 0.001} & 0.076 ± 0.003 & 0.013 ± 0.002 & 0.057 ± 0.001 \\
& \textsc{Ent} & 0.007 ± 0.000 & 0.075 ± 0.003 & 0.012 ± 0.002 & \textbf{0.055 ± 0.003} \\
\midrule
\multirow{5}{*}{\texttt{ACC} ratio ($\uparrow$)}
& \textsc{Erm} & \cellcolor{gray!20} 0.992 ± 0.001 & \cellcolor{gray!20} 0.913 ± 0.003 & \cellcolor{gray!20} \textbf{0.987 ± 0.004} & \cellcolor{gray!20} 0.941 ± 0.000 \\
& \textsc{Unl} & 0.993 ± 0.001 & 0.912 ± 0.003 & 0.982 ± 0.003 & 0.938 ± 0.002 \\
& \textsc{Adv} & 0.993 ± 0.000 & 0.916 ± 0.002 & 0.985 ± 0.004 & 0.941 ± 0.003 \\
& \textsc{Dis} & \textbf{0.994 ± 0.001} & \textbf{0.917 ± 0.003} & 0.985 ± 0.002 & 0.942 ± 0.001 \\
& \textsc{Ent} & 0.993 ± 0.000 & \textbf{0.917 ± 0.003} & 0.985 ± 0.002 & \textbf{0.944 ± 0.003} \\
\midrule
\midrule
\multirow{5}{*}{\texttt{PPV} ($\uparrow$)}
& \textsc{Erm} & \cellcolor{gray!20} 0.896 ± 0.002 & \cellcolor{gray!20} 0.808 ± 0.002 & \cellcolor{gray!20} 0.827 ± 0.001 & \cellcolor{gray!20} 0.860 ± 0.002 \\
& \textsc{Unl} & 0.898 ± 0.002 & 0.807 ± 0.002 & 0.826 ± 0.000 & 0.855 ± 0.003 \\
& \textsc{Adv} & \textcolor{biasgreen}{0.905 ± 0.005} & 0.810 ± 0.002 & 0.829 ± 0.000 & 0.861 ± 0.005 \\
& \textsc{Dis} & \textcolor{biasgreen}{0.918 ± 0.003} & \textcolor{biasgreen}{0.818 ± 0.002} & 0.831 ± 0.001 & 0.863 ± 0.005 \\
& \textsc{Ent} & \textcolor{biasgreen}{\textbf{0.925 ± 0.007}} & \textcolor{biasgreen}{\textbf{0.820 ± 0.007}} & \textcolor{biasgreen}{\textbf{0.833 ± 0.001}} & \textcolor{biasgreen}{\textbf{0.872 ± 0.004}} \\
\midrule
\multirow{5}{*}{$\Delta$\texttt{PPV} ($\downarrow$)}
& \textsc{Erm} & \cellcolor{gray!20} 0.056 ± 0.006 & \cellcolor{gray!20} 0.042 ± 0.002 & \cellcolor{gray!20} 0.005 ± 0.001 & \cellcolor{gray!20} 0.159 ± 0.004 \\
& \textsc{Unl} & 0.054 ± 0.008 & 0.038 ± 0.000 & 0.007 ± 0.004 & \textcolor{biasred}{0.165 ± 0.004} \\
& \textsc{Adv} & \textcolor{biasgreen}{0.048 ± 0.006} & 0.047 ± 0.004 & 0.009 ± 0.001 & \textcolor{biasgreen}{0.152 ± 0.007} \\
& \textsc{Dis} & \textcolor{biasgreen}{0.042 ± 0.003} & 0.039 ± 0.004 & 0.005 ± 0.001 & \textcolor{biasgreen}{0.150 ± 0.006} \\
& \textsc{Ent} & \textcolor{biasgreen}{\textbf{0.031 ± 0.005}} & \textbf{0.037 ± 0.009} & \textbf{0.003 ± 0.001} & \textcolor{biasgreen}{\textbf{0.135 ± 0.004}} \\
\midrule
\multirow{5}{*}{\texttt{PPV} ratio ($\uparrow$)}
& \textsc{Erm} & \cellcolor{gray!20} 0.939 ± 0.007 & \cellcolor{gray!20} 0.948 ± 0.002 & \cellcolor{gray!20} 0.994 ± 0.002 & \cellcolor{gray!20} 0.818 ± 0.004 \\
& \textsc{Unl} & 0.941 ± 0.008 & 0.953 ± 0.001 & 0.992 ± 0.005 & \textcolor{biasred}{0.810 ± 0.005} \\
& \textsc{Adv} & \textcolor{biasgreen}{0.948 ± 0.006} & 0.943 ± 0.005 & 0.989 ± 0.002 & \textcolor{biasgreen}{0.825 ± 0.008} \\
& \textsc{Dis} & \textcolor{biasgreen}{0.955 ± 0.003} & 0.953 ± 0.004 & 0.994 ± 0.001 & \textcolor{biasgreen}{0.829 ± 0.008} \\
& \textsc{Ent} & \textcolor{biasgreen}{\textbf{0.967 ± 0.005}} & \textcolor{biasgreen}{\textbf{0.955 ± 0.011}} & \textbf{0.996 ± 0.001} & \textcolor{biasgreen}{\textbf{0.847 ± 0.005}} \\
\midrule
\midrule
\multirow{5}{*}{\texttt{TPR} ($\uparrow$)}
& \textsc{Erm} & \cellcolor{gray!20} 0.977 ± 0.001 & \cellcolor{gray!20} 0.865 ± 0.002 & \cellcolor{gray!20} 0.827 ± 0.001 & \cellcolor{gray!20} 0.944 ± 0.001 \\
& \textsc{Unl} & 0.977 ± 0.000 & 0.860 ± 0.001 & 0.826 ± 0.000 & 0.943 ± 0.001 \\
& \textsc{Adv} & \textcolor{biasgreen}{0.989 ± 0.001} & \textcolor{biasgreen}{0.871 ± 0.002} & 0.828 ± 0.001 & 0.947 ± 0.001 \\
& \textsc{Dis} & \textcolor{biasgreen}{0.989 ± 0.001} & \textcolor{biasgreen}{0.873 ± 0.003} & 0.831 ± 0.001 & \textcolor{biasgreen}{0.950 ± 0.001} \\
& \textsc{Ent} & \textcolor{biasgreen}{\textbf{0.990 ± 0.000}} & \textcolor{biasgreen}{\textbf{0.875 ± 0.003}} & \textbf{0.832 ± 0.001} & \textcolor{biasgreen}{\textbf{0.953 ± 0.003}} \\
\midrule
\multirow{5}{*}{$\Delta$\texttt{TPR} ($\downarrow$)}
& \textsc{Erm} & \cellcolor{gray!20} 0.009 ± 0.002 & \cellcolor{gray!20} 0.009 ± 0.007 & \cellcolor{gray!20} 0.008 ± 0.005 & \cellcolor{gray!20} 0.067 ± 0.005 \\
& \textsc{Unl} & 0.009 ± 0.004 & \textbf{0.004 ± 0.002} & 0.007 ± 0.003 & 0.066 ± 0.010 \\
& \textsc{Adv} & \textcolor{biasgreen}{\textbf{0.002 ± 0.001}} & 0.013 ± 0.006 & \textcolor{biasred}{0.014 ± 0.006} & \textcolor{biasgreen}{0.057 ± 0.010} \\
& \textsc{Dis} & \textcolor{biasgreen}{0.003 ± 0.001} & 0.011 ± 0.007 & \textcolor{biasred}{0.015 ± 0.003} & \textcolor{biasgreen}{\textbf{0.049 ± 0.009}} \\
& \textsc{Ent} & \textcolor{biasgreen}{0.003 ± 0.002} & 0.008 ± 0.002 & \textcolor{biasgreen}{\textbf{0.002 ± 0.002}} & \textcolor{biasgreen}{0.054 ± 0.013} \\
\midrule
\multirow{5}{*}{\texttt{TPR} ratio ($\uparrow$)}
& \textsc{Erm} & \cellcolor{gray!20} 0.991 ± 0.002 & \cellcolor{gray!20} 0.989 ± 0.008 & \cellcolor{gray!20} 0.990 ± 0.006 & \cellcolor{gray!20} 0.928 ± 0.006 \\
& \textsc{Unl} & 0.990 ± 0.004 & \textcolor{biasgreen}{\textbf{0.995 ± 0.002}} & 0.991 ± 0.004 & 0.929 ± 0.010 \\
& \textsc{Adv} & \textcolor{biasgreen}{\textbf{0.997 ± 0.001}} & 0.985 ± 0.007 & \textcolor{biasred}{0.982 ± 0.007} & \textcolor{biasgreen}{0.939 ± 0.011} \\
& \textsc{Dis} & \textcolor{biasgreen}{\textbf{0.997 ± 0.001}} & 0.988 ± 0.008 & \textcolor{biasred}{0.981 ± 0.004} & \textcolor{biasgreen}{\textbf{0.947 ± 0.010}} \\
& \textsc{Ent} & \textcolor{biasgreen}{0.997 ± 0.002} & 0.990 ± 0.003 & \textcolor{biasgreen}{\textbf{0.997 ± 0.002}} & \textcolor{biasgreen}{0.943 ± 0.014} \\
\midrule
\midrule
\multirow{5}{*}{\texttt{FPR} ($\downarrow$)}
& \textsc{Erm} & \cellcolor{gray!20} 0.023 ± 0.001 & \cellcolor{gray!20} 0.135 ± 0.002 & \cellcolor{gray!20} 0.173 ± 0.001 & \cellcolor{gray!20} 0.056 ± 0.001 \\
& \textsc{Unl} & 0.023 ± 0.000 & 0.140 ± 0.001 & 0.174 ± 0.000 & 0.057 ± 0.001 \\
& \textsc{Adv} & \textcolor{biasgreen}{0.011 ± 0.001} & \textcolor{biasgreen}{0.129 ± 0.002} & 0.172 ± 0.001 & 0.053 ± 0.001 \\
& \textsc{Dis} & \textcolor{biasgreen}{0.011 ± 0.001} & \textcolor{biasgreen}{0.127 ± 0.003} & 0.169 ± 0.001 & \textcolor{biasgreen}{0.050 ± 0.001} \\
& \textsc{Ent} & \textcolor{biasgreen}{\textbf{0.010 ± 0.000}} & \textcolor{biasgreen}{\textbf{0.125 ± 0.003}} & \textbf{0.168 ± 0.001} & \textcolor{biasgreen}{\textbf{0.047 ± 0.003}} \\
\midrule
\multirow{5}{*}{$\Delta$\texttt{FPR} ($\downarrow$)}
& \textsc{Erm} & \cellcolor{gray!20} 0.009 ± 0.002 & \cellcolor{gray!20} 0.009 ± 0.007 & \cellcolor{gray!20} 0.008 ± 0.005 & \cellcolor{gray!20} 0.067 ± 0.005 \\
& \textsc{Unl} & 0.009 ± 0.004 & \textbf{0.004 ± 0.002} & 0.007 ± 0.003 & 0.066 ± 0.010 \\
& \textsc{Adv} & \textcolor{biasgreen}{\textbf{0.002 ± 0.001}} & 0.013 ± 0.006 & \textcolor{biasred}{0.014 ± 0.006} & \textcolor{biasgreen}{0.057 ± 0.010} \\
& \textsc{Dis} & \textcolor{biasgreen}{0.003 ± 0.001} & 0.011 ± 0.007 & \textcolor{biasred}{0.015 ± 0.003} & \textcolor{biasgreen}{\textbf{0.049 ± 0.009}} \\
& \textsc{Ent} & \textcolor{biasgreen}{0.003 ± 0.002} & 0.008 ± 0.002 & \textcolor{biasgreen}{\textbf{0.002 ± 0.002}} & \textcolor{biasgreen}{0.054 ± 0.013} \\
\midrule
\multirow{5}{*}{\texttt{FPR} ratio ($\uparrow$)}
& \textsc{Erm} & \cellcolor{gray!20} 0.713 ± 0.043 & \cellcolor{gray!20} 0.944 ± 0.043 & \cellcolor{gray!20} 0.966 ± 0.020 & \cellcolor{gray!20} 0.504 ± 0.015 \\
& \textsc{Unl} & 0.714 ± 0.109 & \textcolor{biasgreen}{\textbf{0.976 ± 0.011}} & 0.968 ± 0.013 & \textcolor{biasgreen}{0.521 ± 0.033} \\
& \textsc{Adv} & \textcolor{biasgreen}{\textbf{0.822 ± 0.029}} & \textcolor{biasred}{0.917 ± 0.041} & \textcolor{biasred}{0.940 ± 0.025} & \textcolor{biasgreen}{0.541 ± 0.046} \\
& \textsc{Dis} & \textcolor{biasgreen}{0.770 ± 0.074} & \textcolor{biasred}{0.931 ± 0.044} & \textcolor{biasred}{0.931 ± 0.014} & \textcolor{biasgreen}{\textbf{0.568 ± 0.049}} \\
& \textsc{Ent} & \textcolor{biasgreen}{0.794 ± 0.102} & 0.946 ± 0.015 & \textcolor{biasgreen}{\textbf{0.990 ± 0.008}} & \textcolor{biasgreen}{0.540 ± 0.049} \\
\bottomrule
\end{tabular}
}
\caption{Performance evaluation of fine-tuning strategies on CelebA dataset, averaged over three random seeds. The best values are shown in \textbf{bold}. The scores that improve upon \textsc{Erm} by more than 0.005 are highlighted in \textcolor{biasgreen}{green}, while degradations greater than 0.005 are shown in \textcolor{biasred}{red}. Changes within ±0.005 of the \textsc{Erm} score are considered insignificant.}
\label{tab:performance-celeba-2}
\end{table*}

\begin{table*}[ht]
\centering
\resizebox{0.6\linewidth}{!}{
\begin{tabular}{llcccccc}
\toprule
\textbf{Metric} & \textbf{Method} & \multicolumn{3}{c}{\textbf{Avg. Score}} & \multicolumn{3}{c}{\textbf{Avg. Rank} ($\downarrow$)} \\
\cmidrule(lr){3-5}
\cmidrule(lr){6-8}
& & Overall ($\uparrow$) & Diff. ($\downarrow$) & Ratio ($\uparrow$) & Overall ($\uparrow$) & Diff. ($\downarrow$) & Ratio ($\uparrow$) \\
\midrule
\multirow{5}{*}{\texttt{AUROC}}
& \textsc{Erm} & \cellcolor{gray!20}0.968 & \cellcolor{gray!20}0.015 & \cellcolor{gray!20}0.985 & \cellcolor{gray!20}3.77 & \cellcolor{gray!20}3.08 & \cellcolor{gray!20}3.38 \\
& \textsc{Unl} & 0.968 & 0.015 & 0.984 & 4.23 & 3.54 & 3.46 \\
& \textsc{Adv} & 0.970 & 0.014 & 0.986 & 2.62 & 2.46 & 2.54 \\
& \textsc{Dis} & \textbf{0.972} & \textbf{0.013} & \textbf{0.987} & \textbf{1.23} & \textbf{1.92} & \textbf{1.92} \\
& \textsc{Ent} & \textbf{0.972} & \textbf{0.013} & 0.986 & 2.00 & 2.31 & 2.38 \\
\midrule
\multirow{5}{*}{\texttt{AUPRC}}
& \textsc{Erm} & \cellcolor{gray!20}0.895 & \cellcolor{gray!20}0.101 & \cellcolor{gray!20}0.886 & \cellcolor{gray!20}3.92 & \cellcolor{gray!20}3.77 & \cellcolor{gray!20}3.77 \\
& \textsc{Unl} & 0.895 & 0.098 & 0.887 & 4.46 & 3.85 & 3.92 \\
& \textsc{Adv} & \textcolor{biasgreen}{0.912} & \textcolor{biasgreen}{\textbf{0.077}} & \textcolor{biasgreen}{0.914} & 2.54 & 2.38 & 2.38 \\
& \textsc{Dis} & \textcolor{biasgreen}{\textbf{0.915}} & \textcolor{biasgreen}{0.080} & \textcolor{biasgreen}{0.911} & \textbf{1.31} & 2.31 & 2.31 \\
& \textsc{Ent} & \textcolor{biasgreen}{0.913} & \textcolor{biasgreen}{\textbf{0.077}} & \textcolor{biasgreen}{\textbf{0.916}} & 2.08 & \textbf{2.15} & \textbf{2.15} \\
\midrule
\multirow{5}{*}{\texttt{ACC}}
& \textsc{Erm} & \cellcolor{gray!20}0.888 & \cellcolor{gray!20}0.044 & \cellcolor{gray!20}0.949 & \cellcolor{gray!20}3.92 & \cellcolor{gray!20}3.23 & \cellcolor{gray!20}3.31 \\
& \textsc{Unl} & 0.886 & 0.046 & 0.946 & 4.62 & 4.23 & 4.23 \\
& \textsc{Adv} & 0.888 & 0.044 & 0.949 & 3.38 & 2.77 & 2.77 \\
& \textsc{Dis} & \textcolor{biasgreen}{0.894} & 0.040 & \textcolor{biasgreen}{0.955} & 1.85 & 2.15 & 1.77 \\
& \textsc{Ent} & \textcolor{biasgreen}{\textbf{0.897}} & \textcolor{biasgreen}{\textbf{0.038}} & \textcolor{biasgreen}{\textbf{0.956}} & \textbf{1.00} & \textbf{1.46} & \textbf{1.38} \\
\midrule
\multirow{5}{*}{\texttt{PPV}}
& \textsc{Erm} & \cellcolor{gray!20}0.837 & \cellcolor{gray!20}0.065 & \cellcolor{gray!20}0.919 & \cellcolor{gray!20}3.92 & \cellcolor{gray!20}3.62 & \cellcolor{gray!20}3.62 \\
& \textsc{Unl} & 0.835 & 0.065 & 0.917 & 4.54 & 3.38 & 3.46 \\
& \textsc{Adv} & 0.840 & 0.063 & 0.922 & 3.46 & 3.31 & 3.23 \\
& \textsc{Dis} & \textcolor{biasgreen}{0.854} & \textcolor{biasgreen}{\textbf{0.052}} & \textcolor{biasgreen}{0.936} & 1.92 & 2.23 & \textbf{2.00} \\
& \textsc{Ent} & \textcolor{biasgreen}{\textbf{0.863}} & \textcolor{biasgreen}{\textbf{0.052}} & \textcolor{biasgreen}{\textbf{0.937}} & \textbf{1.00} & \textbf{2.15} & 2.15 \\
\midrule
\multirow{5}{*}{\texttt{TPR}}
& \textsc{Erm} & \cellcolor{gray!20}0.881 & \cellcolor{gray!20}0.055 & \cellcolor{gray!20}0.935 & \cellcolor{gray!20}4.00 & \cellcolor{gray!20}3.54 & \cellcolor{gray!20}3.62 \\
& \textsc{Unl} & 0.881 & 0.051 & 0.938 & 4.54 & 3.54 & 3.69 \\
& \textsc{Adv} & \textcolor{biasgreen}{0.895} & \textcolor{biasgreen}{0.044} & \textcolor{biasgreen}{0.949} & 2.92 & 3.00 & 3.08 \\
& \textsc{Dis} & \textcolor{biasgreen}{0.898} & \textcolor{biasgreen}{0.045} & \textcolor{biasgreen}{0.949} & 1.85 & 2.31 & 2.23 \\
& \textsc{Ent} & \textcolor{biasgreen}{\textbf{0.904}} & \textcolor{biasgreen}{\textbf{0.042}} & \textcolor{biasgreen}{\textbf{0.952}} & \textbf{1.23} & \textbf{1.85} & \textbf{1.77} \\
\midrule
\multirow{5}{*}{\texttt{FPR}}
& \textsc{Erm} & \cellcolor{gray!20}0.071 & \cellcolor{gray!20}0.032 & \cellcolor{gray!20}\textbf{0.671} & \cellcolor{gray!20}3.92 & \cellcolor{gray!20}3.08 & \cellcolor{gray!20}\textbf{2.69} \\
& \textsc{Unl} & 0.073 & 0.033 & \textcolor{biasred}{0.659} & 4.54 & 3.69 & 3.00 \\
& \textsc{Adv} & 0.067 & 0.030 & \textcolor{biasred}{0.631} & 3.00 & 3.38 & 3.23 \\
& \textsc{Dis} & \textcolor{biasgreen}{0.063} & 0.028 & \textcolor{biasred}{0.664} & 1.77 & 2.38 & 3.15 \\
& \textsc{Ent} & \textcolor{biasgreen}{\textbf{0.061}} & \textbf{0.027} & \textcolor{biasred}{0.637} & \textbf{1.23} & \textbf{1.62} & 2.92 \\
\bottomrule
\end{tabular}
}
\caption{Average utility and fairness scores and ranks for each method, aggregated over all experiments. The best values are shown in \textbf{bold}. Average scores that improve upon \textsc{Erm} by more than 0.005 are highlighted in \textcolor{biasgreen}{green}, while degradations greater than 0.005 are shown in \textcolor{biasred}{red}. Changes within ±0.005 of the \textsc{Erm} score are considered insignificant.}
\label{tab:average-table}
\end{table*}

\begin{table*}[ht]
\centering
\resizebox{0.7\linewidth}{!}{
\begin{tabular}{lrcccccc}
\toprule
\textbf{Method} & \textbf{Value} & \multicolumn{6}{c}{\textbf{UTKFace (Age $\times$ Race)}} \\
\cmidrule(lr){3-8}
& & \texttt{ROC} ($\uparrow$) & $\Delta$\texttt{ROC} ($\downarrow$) & \texttt{ROC} ratio ($\uparrow$) & \texttt{PR} ($\uparrow$) & $\Delta$\texttt{PR} ($\downarrow$) & \texttt{PR} ratio ($\uparrow$) \\
\midrule
\multirow{5}{*}{\textsc{Unl} ($\lambda_\textsc{Unl}$)}
& - 0.25 & 0.882 ± 0.008 & 0.088 ± 0.006 & 0.904 ± 0.007 & 0.531 ± 0.022 & 0.212 ± 0.020 & 0.657 ± 0.023 \\
& - 0.50 & 0.938 ± 0.001 & 0.073 ± 0.007 & 0.924 ± 0.007 & 0.694 ± 0.032 & 0.225 ± 0.005 & 0.722 ± 0.012 \\
& - 0.75 & \textbf{0.949 ± 0.002} & \textbf{0.048 ± 0.007} & \textbf{0.950 ± 0.008} & \textbf{0.744 ± 0.015} & 0.234 ± 0.044 & 0.729 ± 0.049 \\
& - 1.00 & 0.945 ± 0.000 & 0.049 ± 0.007 & 0.948 ± 0.007 & 0.707 ± 0.013 & \textbf{0.190 ± 0.010} & \textbf{0.768 ± 0.007} \\
& - 1.25 & 0.938 ± 0.007 & 0.056 ± 0.005 & 0.942 ± 0.006 & 0.695 ± 0.003 & 0.218 ± 0.029 & 0.736 ± 0.034 \\
\midrule
\multirow{5}{*}{\textsc{Adv} ($\lambda_\textsc{Adv}$)}
& 0.01 & 0.949 ± 0.001 & 0.033 ± 0.005 & 0.965 ± 0.005 & \textbf{0.818 ± 0.010} & 0.191 ± 0.020 & 0.784 ± 0.025 \\
& 0.05 & 0.949 ± 0.001 & \textbf{0.033 ± 0.004} & \textbf{0.966 ± 0.005} & 0.816 ± 0.010 & 0.193 ± 0.022 & 0.782 ± 0.027 \\
& 0.10 & \textbf{0.950 ± 0.000} & 0.038 ± 0.002 & 0.960 ± 0.002 & 0.814 ± 0.002 & \textbf{0.105 ± 0.009} & \textbf{0.881 ± 0.008} \\
& 0.50 & 0.948 ± 0.001 & \textbf{0.033 ± 0.004} & 0.965 ± 0.005 & 0.816 ± 0.010 & 0.186 ± 0.022 & 0.789 ± 0.028 \\
& 1.00 & 0.949 ± 0.001 & \textbf{0.033 ± 0.004} & 0.965 ± 0.005 & 0.814 ± 0.012 & 0.187 ± 0.020 & 0.788 ± 0.026 \\
\midrule
\multirow{5}{*}{\textsc{Dis} ($\lambda_\textsc{Dis}$)}
& 0.01 & \textbf{0.952 ± 0.000} & \textbf{0.032 ± 0.004} & 0.966 ± 0.005 & \textbf{0.836 ± 0.011} & 0.190 ± 0.018 & 0.786 ± 0.023 \\
& 0.05 & \textbf{0.952 ± 0.000} & \textbf{0.032 ± 0.004} & 0.966 ± 0.005 & 0.835 ± 0.011 & 0.190 ± 0.021 & 0.785 ± 0.026 \\
& 0.10 & 0.951 ± 0.001 & 0.042 ± 0.006 & 0.956 ± 0.006 & 0.815 ± 0.014 & \textbf{0.164 ± 0.046} & \textbf{0.817 ± 0.050} \\
& 0.50 & \textbf{0.952 ± 0.000} & 0.032 ± 0.005 & \textbf{0.967 ± 0.005} & 0.835 ± 0.011 & 0.192 ± 0.019 & 0.783 ± 0.025 \\
& 1.00 & \textbf{0.952 ± 0.000} & 0.032 ± 0.005 & 0.966 ± 0.005 & \textbf{0.836 ± 0.011} & 0.192 ± 0.017 & 0.784 ± 0.022 \\
\midrule
\multirow{5}{*}{\textsc{Ent} ($\lambda_\textsc{Ent}$)}
& 0.10 & \textbf{0.955 ± 0.001} & 0.031 ± 0.005 & 0.968 ± 0.005 & 0.817 ± 0.012 & 0.223 ± 0.029 & 0.752 ± 0.029 \\
& 0.50 & 0.953 ± 0.002 & \textbf{0.030 ± 0.005} & \textbf{0.969 ± 0.005} & 0.813 ± 0.012 & 0.215 ± 0.023 & 0.758 ± 0.025 \\
& 1.00 & 0.947 ± 0.007 & 0.046 ± 0.008 & 0.953 ± 0.008 & \textbf{0.822 ± 0.006} & \textbf{0.144 ± 0.038} & \textbf{0.840 ± 0.040} \\
& 5.00 & 0.949 ± 0.003 & 0.033 ± 0.007 & 0.965 ± 0.007 & 0.800 ± 0.006 & 0.219 ± 0.016 & 0.753 ± 0.026 \\
& 10.00 & 0.952 ± 0.001 & 0.037 ± 0.005 & 0.962 ± 0.005 & 0.805 ± 0.003 & 0.201 ± 0.008 & 0.774 ± 0.009 \\
\bottomrule
\end{tabular}
}
\caption{Performance evaluation of fine-tuning strategies on UTKFace dataset, averaged over three random seeds (for varying method-specific hyperparameter values). The best values for each method are shown in \textbf{bold}.}
\label{tab:hyper-auc-performance-utk-age}
\end{table*}

\begin{table*}[ht]
\centering
\resizebox{0.7\linewidth}{!}{
\begin{tabular}{lrcccccc}
\toprule
\textbf{Method} & \textbf{Value} & \multicolumn{6}{c}{\textbf{UTKFace (Gender $\times$ Race)}} \\
\cmidrule(lr){3-8}
& & \texttt{ROC} ($\uparrow$) & $\Delta$\texttt{ROC} ($\downarrow$) & \texttt{ROC} ratio ($\uparrow$) & \texttt{PR} ($\uparrow$) & $\Delta$\texttt{PR} ($\downarrow$) & \texttt{PR} ratio ($\uparrow$) \\
\midrule
\multirow{5}{*}{\textsc{Unl} ($\lambda_\textsc{Unl}$)}
& - 0.25 & 0.915 ± 0.005 & 0.083 ± 0.005 & 0.911 ± 0.005 & 0.907 ± 0.003 & 0.066 ± 0.003 & 0.929 ± 0.003 \\
& - 0.50 & 0.975 ± 0.003 & 0.035 ± 0.002 & 0.964 ± 0.003 & 0.968 ± 0.004 & \textbf{0.027 ± 0.008} & \textbf{0.972 ± 0.008} \\
& - 0.75 & 0.984 ± 0.002 & 0.030 ± 0.005 & 0.970 ± 0.005 & 0.977 ± 0.004 & 0.039 ± 0.014 & 0.960 ± 0.015 \\
& - 1.00 & \textbf{0.985 ± 0.001} & 0.026 ± 0.002 & 0.974 ± 0.002 & \textbf{0.978 ± 0.003} & 0.037 ± 0.011 & 0.962 ± 0.012 \\
& - 1.25 & 0.983 ± 0.002 & \textbf{0.025 ± 0.004} & \textbf{0.975 ± 0.004} & 0.975 ± 0.002 & 0.030 ± 0.011 & 0.970 ± 0.012 \\
\midrule
\multirow{5}{*}{\textsc{Adv} ($\lambda_\textsc{Adv}$)}
& 0.01 & \textbf{0.986 ± 0.000} & \textbf{0.026 ± 0.002} & 0.973 ± 0.002 & \textbf{0.981 ± 0.001} & \textbf{0.025 ± 0.004} & \textbf{0.975 ± 0.004} \\
& 0.05 & \textbf{0.986 ± 0.000} & 0.027 ± 0.002 & 0.973 ± 0.002 & \textbf{0.981 ± 0.001} & 0.026 ± 0.004 & 0.974 ± 0.004 \\
& 0.10 & 0.985 ± 0.001 & \textbf{0.026 ± 0.002} & 0.973 ± 0.002 & 0.979 ± 0.001 & 0.030 ± 0.004 & 0.970 ± 0.004 \\
& 0.50 & \textbf{0.986 ± 0.000} & \textbf{0.026 ± 0.002} & \textbf{0.974 ± 0.002} & \textbf{0.981 ± 0.001} & \textbf{0.025 ± 0.004} & 0.974 ± 0.004 \\
& 1.00 & \textbf{0.986 ± 0.000} & 0.027 ± 0.002 & 0.973 ± 0.002 & \textbf{0.981 ± 0.001} & \textbf{0.025 ± 0.004} & 0.974 ± 0.004 \\
\midrule
\multirow{5}{*}{\textsc{Dis} ($\lambda_\textsc{Dis}$)}
& 0.01 & \textbf{0.989 ± 0.000} & \textbf{0.023 ± 0.005} & \textbf{0.977 ± 0.005} & 0.985 ± 0.001 & \textbf{0.029 ± 0.006} & \textbf{0.971 ± 0.006} \\
& 0.05 & \textbf{0.989 ± 0.000} & \textbf{0.023 ± 0.005} & \textbf{0.977 ± 0.005} & \textbf{0.985 ± 0.000} & \textbf{0.029 ± 0.006} & \textbf{0.971 ± 0.006} \\
& 0.10 & 0.986 ± 0.001 & 0.028 ± 0.001 & 0.972 ± 0.002 & 0.981 ± 0.002 & 0.031 ± 0.008 & 0.969 ± 0.008 \\
& 0.50 & \textbf{0.989 ± 0.000} & \textbf{0.023 ± 0.005} & \textbf{0.977 ± 0.005} & \textbf{0.985 ± 0.000} & \textbf{0.029 ± 0.006} & \textbf{0.971 ± 0.006} \\
& 1.00 & \textbf{0.989 ± 0.000} & \textbf{0.023 ± 0.005} & \textbf{0.977 ± 0.005} & \textbf{0.985 ± 0.000} & \textbf{0.029 ± 0.006} & \textbf{0.971 ± 0.006} \\
\midrule
\multirow{5}{*}{\textsc{Ent} ($\lambda_\textsc{Ent}$)}
& 0.10 & \textbf{0.990 ± 0.000} & \textbf{0.021 ± 0.007} & \textbf{0.979 ± 0.007} & \textbf{0.987 ± 0.001} & 0.029 ± 0.007 & 0.971 ± 0.007 \\
& 0.50 & \textbf{0.990 ± 0.000} & 0.022 ± 0.006 & 0.978 ± 0.006 & \textbf{0.987 ± 0.001} & 0.032 ± 0.003 & 0.968 ± 0.003 \\
& 1.00 & 0.988 ± 0.001 & 0.027 ± 0.001 & 0.973 ± 0.001 & 0.983 ± 0.002 & \textbf{0.028 ± 0.002} & \textbf{0.972 ± 0.002} \\
& 5.00 & 0.988 ± 0.001 & 0.025 ± 0.007 & 0.975 ± 0.007 & 0.985 ± 0.002 & 0.035 ± 0.008 & 0.965 ± 0.008 \\
& 10.00 & 0.984 ± 0.001 & 0.027 ± 0.005 & 0.973 ± 0.005 & 0.981 ± 0.002 & 0.032 ± 0.009 & 0.967 ± 0.009 \\
\bottomrule
\end{tabular}
}
\caption{Performance evaluation of fine-tuning strategies on UTKFace dataset, averaged over three random seeds (for varying method-specific hyperparameter values). The best values for each method are shown in \textbf{bold}.}
\label{tab:hyper-auc-performance-utk-gender}
\end{table*}


\end{document}